# A novel approach to measuring the scope of patent claims based on probabilities obtained from (large) language models

**Author**: Sébastien Ragot

**Professional address**: E. Blum & Co. Ltd., Patent and Trademark Attorneys VSP, Hofwiesenstrasse 349, 8050 Zürich, Switzerland

**Abstract**

This work proposes to measure the scope of a patent claim as the reciprocal of self-information contained in this claim. Self-information is calculated based on a probability of occurrence of the claim, where this probability is obtained from a language model. Grounded in information theory, this approach is based on the assumption that an unlikely concept is more informative than a usual concept, insofar as it is more surprising. In turn, the more surprising the information required to define the claim, the narrower its scope. Seven language models are considered, ranging from simplest models (each word or character has an identical probability) to intermediate models (based on average word or character frequencies), to large language models (LLMs) such as GPT2 and davinci-002. Remarkably, when using the simplest language models to compute the probabilities, the scope becomes proportional to the reciprocal of the number of words or characters involved in the claim, a metric already used in previous works. Application is made to multiple series of patent claims directed to distinct inventions, where each series consists of claims devised to have a gradually decreasing scope. The performance of the language models is then assessed through several ad hoc tests. The LLMs outperform models based on word and character frequencies, which themselves outdo the simplest models based on word or character counts. Interestingly, however, the character count appears to be a more reliable indicator than the word count.

# Table of Contents

# 1. Introduction

Patent language is known to be convoluted. In particular, patent claims can be difficult to interpret, also for intellectual property (IP) stakeholders. The latter must, however, be able to fully understand the scope of patent claims (i.e., their breadth, depth, and reach), if only because it determines their patentability and the extent to which they can be enforced.

Besides its legal consequences, the scope of patent claims also impacts the valuation of patents, hence the need of a measure of the scope.

## 1.1. Antagonistic origins of the complexity of patent language

The scope of a patent's protection is primarily defined by patent claims, which are appended to a detailed description of the invention. Each patent claim is a single-sentence statement that defines the technical features of the invention using legal phrasing. Claims often involve circumlocutions, if not linguistic aerobatics, which makes them difficult to read. The same goes for the detailed description, which elaborates on the claims, hence the odd reputation of patent language.

While certain turns of phrase have a long tradition, the complexity of patent language essentially results from a necessary trade-off between the antagonistic interests of patent players and third parties. On the one hand, patent applicants want to secure the broadest possible scope of protection. On the other hand, they must comply with legal clarity and other patentability requirements, as well as further legal constraints aimed at preserving the interests of third parties. So, the problem faced by patent applicants resembles that of maximization under constraints.

In practice, a patent office examines any patent application filed to ensure that all legal provisions are met before possibly granting a patent. Like a negotiation process, this examination gives rise to multiple interactions (the "patent prosecution") with the applicants or their representatives [1]. Applicants generally try to optimally amend the patent claims under various legal constraints, which further increases the complexity of patent language.

In detail, the readability of patent claims is first impacted by the fact that patent drafters attempt to cover all possible realizations (called "embodiments") of the invention, including variants that competitors might otherwise consider to avoid infringement. All possible embodiments must ideally be covered by the claims and supported by the detailed description of the patent application before filing this application with a patent office. This is desirable as the extent to which the claims can be subsequently amended is restricted by the content of the application as initially filed. That is, it is not possible to add any new subject matter after filing the patent application, see, e.g., Article 123 EPC [2, 3]. Such legal provisions are necessary to preserve the interests of third parties and are strictly enforced by patent offices and courts. Thus, patent drafters must carefully compose the claims from the outset.

The same goes for the detailed description, which elaborates on the claims. The description must describe the invention in sufficient detail so that third parties can carry out the invention, see, e.g., Article 83 EPC [2, 3]. This requirement ensures a fair trade-off: a patent gives its owner an exclusive right to manufacture, use, or sell an invention for a limited period in exchange for publishing the disclosure of the invention [4]. That is, the patent owner is granted an exclusive right, for a limited period, in exchange for sufficient disclosure of the invention.

Still, applicants often endeavor not to disclose details that are not essential to making or practicing embodiments falling within the scope of the claims, particularly where such details allow them to maintain a competitive advantage. The patent drafting process can thus be compared to musical composition, where silences are just as important as the notes played.

Any technical feature claimed or described can later be exploited to restore novelty, should the patent examination process reveal that the broadest claims are not patentable, e.g., not novel or not inventive, see Article 52 of the European Patent Convention (EPC) [2]. Patent applicants then strive to adjust the scope of the broadest claims (to the extent permitted by the contents of the patent application as initially filed) for them to be novel enough to be patentable.

Another legal requirement is clarity, see, e.g., Article 84 EPC [2, 3], which requires the claims to be concise and prohibits any lack of antecedent basis in the claims. That is, the claims must clearly identify the various concepts involved and then consistently refer to such concepts, which often leads to repetitions and convoluted wording. For example, a claim may initially define a *device* containing a *first part* and a *second part*, as well as *attachment means that include one or more screws*. The subsequently claimed features must then unambiguously refer to the concepts defined, like variables in a procedural programming language. And this may give rise to convoluted constructs, such as *the first part is fixed to the second part thanks to at least one of the one or more screws*.

### 1.2. Purpose of the paper

This complexity impacts the intelligibility of the claims, which makes it difficult for IP stakeholders to appreciate their reach. First, it is necessary to clearly understand the scope of a claim (in the sense of its breadth) to determine its patentability. To that aim, patent examiners consider the claims individually, to determine what each claim encompasses and, in turn, assess its novelty and non-obviousness in view of the prior art [3]. The aim is to identify what exactly is patentable, as opposed to what is not. Second, once granted, claims are taken into consideration by patent owners in: (i) deciding whether to keep the patents alive; and (ii) determining which claims may possibly be enforced against potential competitors. Third, litigation is primarily about claims as well, again considered individually. Depending on the requests and allegations of the parties and the case at hand, the courts may have to examine all patentability requirements in respect of any patent claim alleged to be infringed. Only then will it make sense to assess the alleged infringement. Thus, there are multiple needs to precisely understand the scope of patent claims.

Besides the need to understand patent claims, it is furthermore needed to quantify the scope of patent claims, if only to automate comparisons. A sound measure would for instance be particularly useful to legal experts and other IP stakeholders to assess the value of patents for patent transactions; granted patent claims having a broader scope have greater potential, whether in terms of licensing or litigation. Moreover, for freedom-to-operate studies, such a measure may be used together with a relevance score to rank patent documents returned by patent databases in response to user queries. Indeed, freedom to operate is foremost impacted by relevant patents that have the broadest scope. For completeness, automated claim scope calculations may also be used as a quality metric by patent drafters and patent drafting tools. What is sought in all such examples is a mere valuation, i.e., a method to equate the claim scope to a number.

A fundamental question that raises then is how to measure the scope of a patent or a patent claim. Various approaches have been proposed, which mostly rely on bibliometric indicators and claim lengths, e.g., the word count of the very first claim. However, such indicators are either unrelated to the practice of IP players or can be misleading, as discussed in Section 2. Therefore, an object of the present work is to design a more reliable measure of the scope of patent claims.

With this in mind, this work proposes to measure the scope of a patent claim as the reciprocal (i.e., the multiplicative inverse) of self-information contained in that claim. Self-information is calculated from a probability associated with this claim. This probability is an estimate of the probability of occurrence, or completion, of the claim text. It is obtained thanks to a given language model, e.g., a Large Language Model (LLM). This approach is grounded in information theory and surprisal analysis. It is based on the assumption that an unlikely concept is more informative than a usual concept, inasmuch as it is more surprising. In turn, the more information content required to define the claim, the narrower its scope. Results are presented and assessed in Section 4. The probability models are compared in Section 5. Section 6 closes this paper with a few concluding remarks.

Before discussing prior works (Section 2) and their limits in detail, it is necessary to recall the purpose of patent claims and how they contribute to defining the scope of a claimed invention.

## 1.3. Patent claim structure

Two basic categories of patent claims are: (i) the independent claims, which stand on their own; and (ii) the dependent claims, which depend on one or more parent claims and thus include all the limitations of the parent claims.

Independent claims capture the gist of the invention, while dependent claims reflect particular embodiments of the invention. The dependent claims explore additional features, thereby adding complexity to claim features already recited in the claims they depend on. They can notably serve as fallback positions during the patent prosecution.

The patent examination process may for example reveal the existence of a prior art document that anticipates an independent claim, such that this independent claim is effectively not novel. In that case, the patent applicant may fall back on a dependent claim to restore novelty. Actually, any residual novel feature may potentially be exploited to restore novelty, whether extracted from a dependent claim, the detailed description of the invention, or even the accompanying drawings, as initially filed. A patent may eventually be granted for a revised set of claims, in which independent claims are amended to recite this novel feature, provided that the claim amendments are supported by the initial application content (see, e.g., Article 123(2) EPC [2, 3]) and the amended claims satisfy other patentability requirements, starting with legal clarity, industrial application, and inventive step (see, e.g., Article 52 EPC [2, 3]), also called non-obviousness.

As a whole, the claims define the extent of the protection, i.e., which subject matter is protected by a granted patent or sought in a patent application before grant, see, e.g., Article 84 EPC [2, 3]. The claims can accordingly be regarded as a “notice function”, warning third parties of what they must not do to avoid infringement liability.

Let us illustrate this with a simple example of an invention directed to a pencil eraser. A corresponding patent claim may essentially recite an eraser attached to an end of the pencil —a concept that was manifestly novel at the time of the invention [5]. The first claim may for instance read:

> *1. A writing instrument including a pencil and an eraser, wherein the eraser is fixed to the pencil.*

Next, a dependent claim (say, claim 2, depending on claim 1) could be:

*2. The writing instrument according to claim 1, wherein the writing instrument further includes a ferrule, the eraser being fixed to the pencil through the ferrule.*

Further dependent claims may similarly be appended to capture particular embodiments of the invention. For example, dependent claims may explore a preferred type of metal or alloy in the ferrule, and/or the type of ferrule. More generally, dependent claims seek to cover all interesting embodiments, whether horizontally (all alternative embodiments are successively explored) or vertically (each alternative is gradually narrowed down by refining the recited features or introducing new features). Each dependent claim can be regarded as defining a further invention, which falls within the scope of the more general invention defined by its parent claim.

A same patent application may contain several independent patent claims, belonging to different categories. For instance, in the above example, a second independent claim may address a method of fabrication of the pencil eraser, e.g.,

*3. A method of fabricating a writing instrument including a pencil and an eraser, wherein the method comprises fixing the eraser to the pencil.*

Going further, a dependent method claim may read:

*4. The method according to claim 3, wherein, at fixing, the eraser is fixed to the pencil through a ferrule.*

As illustrated above, there are essentially two categories of claims, corresponding to products (i.e., tangible things) and activities (i.e., methods, processes, etc.). Each category may include two types of claims, i.e., the independent and dependent claims. In the above example, claims 1 and 2 are directed to an object, while claims 3 and 4 concern an activity. However, the two independent claims (claims 1 and 3) revolve around the same inventive concept, i.e., attaching an eraser to a pencil, as required to fulfill the condition of unity of the invention, see, e.g., Article 82 EPC [2, 3].

Each independent claim can be regarded as a root node of a tree structure. Successive child nodes (i.e., dependent claims) along a same branch of the tree have a gradually decreasing scope. In the above example, claim 2 is narrower than claim 1 because it involves a ferrule (i.e., a metal sleeve), while claim 1 potentially covers any fixation means. The same goes for claim 4 with respect to claim 3. Siblings on respective branches of a tree capture alternatives, having distinct scopes. For example, a further dependent claim (claim 5, depending on claim 4) may require the eraser to be first inserted in the ferrule on one side, before inserting the pencil on the other side. Conversely, a sibling claim (claim 6, depending on claim 4 but not on claim 5) may address an alternative fabrication process where the pencil is first inserted in the ferrule on one side, before inserting the eraser on the other side.

Of course, the reality of contemporary patent claims is often much more complex. Still, the above examples illustrate that each claim has a respective scope. And this tells that the scope of a patent should primarily be defined on a claim level.

# 2. Prior works and limitations of prior approaches

## 2.1. Prior works on patent scope

Several patent scope metrics have been proposed. In particular, bibliometric indicators have been used extensively in the past, as for instance noted in [6, 7, 8, 9, 10, 11]. Such indicators typically involve the number of patent classes, the number of backward and forward citations, the count of inventors and assignees, and the number of claims.

Bibliometric indicators have long puzzled patent practitioners as such indicators only play a secondary role in practice: IP stakeholders (including patent offices and courts) primarily rely on patent claims. Several papers have consequently noted the lack of practical relevance of such indicators and scholars proposed instead to rely on the length of the claims to measure the scope of the patent, as for instance illustrated in the works of Jansen [7], Osenga [8], Yokada et al. [9], Marco et al. [10], as well as Kuhn and Thompson [11]. Such works focus on independent claims, if not the very first claim. They mostly rely on word counts, i.e., the number of words contained in a claim, subject to, e.g., Yokada et al., who use character counts because of the specificities of the Japanese language [9].

The above references provide various justifications for the use of the claimed length, whether measured through the word count or character count. For example, Kuhn and Thompson [11] assert that this approach is grounded both in patent law and in the practices of patent attorneys; it would even be widely recognized by practitioners, according to a comment by Quinn [12]. This, however, must be nuanced as several patent practitioners have also challenged the role played by the word counts [13]. The word and character counts can indeed be misleading and the extent to which they make it possible to discriminate between claims is questionable, as discussed below.

## 2.2. Drawbacks of scope measures based on the word or character count of a claim

Consider a draft claim (claim 1) that initially involves three technical features—call them A, B, and C. Assume further that the patent drafter ultimately decides to remove one of these features (e.g., feature C) and place it in a dependent claim (e.g., claim 2). This increases the scope and reduces the word count of claim 1. Conversely, the patent drafter may decide to add a further feature (feature D) to features A, B, and C, to reinforce novelty before filing the patent application at a patent office. This narrows down the scope of claim 1, now reciting A, B, C, and D, which also normally increases the word count. In that sense, the word count indeed varies according to the number of claimed features, hence a correlation between the actual scope and the word count.

However, the word character count can also be misleading as numerous claim variations (of various lengths) may capture the same invention with the exact same scope. Indeed, patent drafters necessarily take a subjective approach to inventions. For instance, bringing together hundreds of practitioners and tasking them with drafting an independent patent claim for the same invention rarely results in any two perfectly identical claims, as correctors of the European Qualifying Examination correctors may confirm [14]. Rather, the resulting claims deviate from each other, whether in terms of terminologies or phrasing. This translates into a dispersion of the word and character counts. To that extent, such counts cannot be accurate indicators of the claim scope.

Going back to the example of the pencil eraser, the very first claim of a patent application could read:

> *1. A writing instrument including a pencil and an eraser, wherein the eraser is fixed to an end of the pencil.*

The above patent claim is 20 words long and includes 106 characters, not counting the first three characters (“1. ”) and the full stop. This claim is concise as it contains little superfluous language. Note, while stop words are often considered insignificant in natural language processing (NLP), they often play a decisive role in patent claims, if only because of the relations they imply. That said, a claim can always be formulated using a more or less concise wording. For example, the above claim 1 may be rephrased as

> *1a. Writing instrument including a pencil and an eraser, the latter fixed to an end of the pencil.*

Claim 1a counts 17 words and 93 characters only. Conversely, this claim may also be rewritten less concisely, e.g.,

> *1b. A writing instrument, wherein the writing instrument includes both a pencil and an eraser, and wherein the eraser is fixed to an end of the pencil.*

Claim 1b consists of 26 words and 146 characters. Going further, one realizes that hundreds to thousands of claim variants can be obtained by merely playing with the word arrangement and stop words. The word and character counts can therefore vary considerably. For instance, reasonable claim variants to the pencil eraser invention may contain between 15 to 28 words, and between 81 and 148 characters. Since such claims address the exact same invention, an ideal model should predict the exact same scope value. Far from it, the claim scope is found to vary from $1/15 \approx 0.066$ to $1/28 \approx 0.036$ when calculated as the reciprocal of the word count or from $1/81 \approx 0.012$ to $1/148 \approx 0.007$ when using the character count in place of the word count.

Thus, it is legitimate to question indicators based on the claim length, whether measured as a word or character count. Another reason to do so is that such indicators cannot discriminate between two claims having an identical length. For example, a *computing device* (synonymous with *computer*) necessarily has a broader scope than a *quantum computer*, even though the two statements have identical word counts and character counts.

### 2.3. Need for improvement

Common sense tells a careful assessment of the scope of the claims must necessarily take semantics into account. I.e., assessing the reach of a claim requires an analysis of the meaning of the recited words.

Various types of semantic analyses of patent claims have been reported. For example, Wittfoth has proposed to measure the patent scope thanks to a normalized indicator obtained through a semantic analysis of the patent claims [6]. As another example, Lee et al. use a semantic patent claim analysis based on dependency relations, whereby hierarchical keyword vectors are used to represent the dependency relations among the claim elements. A tree-matching algorithm is used to compare such claim elements [15]. Further semantic studies are reported in the review made by Abbas et al. [16].

However, semantic analyses are complex and can be time-consuming. Also, there is, to date, no universally accepted approach for the semantic analysis of patent claims. On the contrary, language models have become established in translation and automatic text generation. It is therefore desirable to explore their potential for automatic evaluation of the scope of claims.

# 3. Defining the scope of patent claims based on probabilities obtained from language models

The present approach is based on probabilities associated with claims, where such probabilities can be regarded as probabilities of occurrence or completion, which can be determined thanks to a language model. A language model is a probabilistic model of a natural language, which can be used to generate text based on probabilities of textual signs. Conversely, a language model can also be used to calculate a probability of occurrence of a claim based on probabilities of such signs.

Large language models (LLMs) are advanced computational models, which are typically trained on large text corpora. They typically combine different architectures of neural networks, e.g., feedforward neural networks and transformers. Beyond LLMs, other language models can be devised, starting with simplistic models, where text elements are generated randomly or based on average frequencies.

In principle, the probability of occurrence of a claim may be directly used to measure the scope of this claim. However, the probabilities obtained from language models may easily differ by several orders of magnitude, even for claims addressing the same invention with the exact same scope, should the claims use a slightly different wording. This advocates using a logarithmic scale, hence the proposition to rely on self-information, a well-accepted concept.

## 3.1. Claim scope as the reciprocal of self-information of a claim

Self-information (also called surprisal, information content, or Shannon information) is defined as the negative log-probability. That is, given a claim $C$, and the probability $p(C)$ of occurrence of this claim, the associated self-information $I(C)$ is defined as:

$$I(C) = -\log\big(p(C)\big) = \log(1/p(C)). \qquad \text{Eq. (1)}$$

While $p(C)$ is represented by a real number in the interval [0, 1], $I(C)$ is represented by a real number in [0, ∞]. The smaller the probability, the larger the surprisal and, thus, the larger the self-information. This definition complies with Shannon's definition of self-information, which was devised to meet several axioms:

(i) An event with probability 1 is fully unsurprising and therefore yields no information ($I = 0$);

(ii) The less probable an event, the more surprising it is and the more information it contains; and

(iii) The total amount of information of two independent events is equal to the sum of the self-information obtained for the individual events, hence the logarithmic dependency.

The concept of self-information is closely related to entropy, which can generally be defined as the expected value of self-information of a random variable. Unlike the entropy, however, self-information is about the information content of a particular event [17]. Thus, self-information is the quantity of interest here.

Eventually, the scope $S(C)$ of a claim can be defined as the reciprocal of $I(C)$, whereby

$$S(C) = 1/I(C) = -1/\log\big(p(C)\big) = 1/\log(1/p(C)). \qquad \text{Eq. (2)}$$

According to this formulation, $S \longrightarrow 0$ as $p \longrightarrow 0$ and $I \longrightarrow \infty$, while $S \longrightarrow \infty$ as $p \longrightarrow 1$ and $I \longrightarrow 0$. In other words, a claim having a small value of self-information (meaning its definition requires little information) has a broad scope, whereas a claim requiring a lot of information to define it has a narrow scope [18, 19]. That is, while $I(C)$ measures the information content of claim $C$, the scope $S(C)$ measures the breadth of claim $C$ as determined by $I(C)$. Self-information can thus be compared to a number density (i.e., the amount of information per claim), in which case the scope $S(C)$ can be compared to a volume. The higher the density of the claim (i.e., the more information it contains), the smaller its volume (or scope), and the lower its density, the larger its volume. Unlike the probability $p$, the quantities $I$ and $S$ are not bounded, consistently with the natural perception one has of the amount of information contained in a claim involving an increasingly large number of words.

The scope as defined by Eq. (2) is based on the entire text of a claim and not on the gist of the invention—it is in fact agnostic to the gist of the invention. This definition is consistent with the so-called peripheral definition of the patent scope [20, 21]. Under this definition, claims define outer boundaries of protection, like linguistic "fence posts", which delimit the patentee's exclusive territory. An advantage of this approach is that the scope of protection will, in principle, be clear to a third party who reads (and understands) the claims, thereby providing legal certainty.

Another approach to interpreting the scope of the claims is the so-called central definition, according to which the scope of protection is determined in accordance with the gist (or "inventive idea") of the invention, i.e., the principle underlying the invention. This approach is not strictly bound by the actual wording of the claims.

In practice, patent offices and courts often adopt intermediate positions. The patent systems of continental Europe are usually considered more central than those of the United Kingdom, the United

States, and Japan's [21]. For example, under the EPC, the invention is defined by the claims (Article 84 EPC), which are also used to determine the patent's scope of protection (Article 69 EPC). Still, the claims must be interpreted in light of the description and the drawings (Article 69 EPC), whereby the claim language must be interpreted in the context in which it is used. That said, the claims play a determinative part in deciding on the patent scope. Also, the need for legal certainty globally urges the courts towards a peripheral interpretation of the claims, something that calls for greater caution when drafting patent claims.

Since, in the present work, the scope is defined on a claim level, it only reflects the scope of a given claim, not the patent. Any claim may possibly be considered, starting with independent claims. If needed, various schemes can be contemplated to determine the global scope of the patent. For example, the patent scope may be defined based on the scope of the broadest claim only, as the latter sets an upper limit to the patent scope. By definition, the broadest claim is one of the independent claims, not necessarily the first claim. In practice, the broadest claim is often directed to an object or product (a "product claim"), if any, as this claim potentially covers any use of that product. In variants, the scope of a patent can be defined as the average scope of all the independent claims.

However, as noted earlier, the scope should primarily be appreciated with respect to each claim, considered individually, consistently with patent prosecution and litigation procedures. Thus, in the present work, the scope is defined and investigated on a claim level.

## 3.2. Probability of occurrence of a claim

### 3.2.1. Formulation

The above subsection defines the scope $S(C)$ of a patent claim $C$ as the reciprocal of $I(C)$, which itself depends logarithmically on the probability $p(C)$ of occurrence (or completion) of a given claim. There remains to define practical schemes to obtain $p(C)$. To that end, use can be made of a language model.

This model can for instance be an LLM, such as GPT-*n* [22]. In some cases, LLMs can be exploited to formulate the probability of occurrence of a given sentence, e.g., as a chain rule of conditional probabilities, i.e., a product of probability terms, where each term is conditioned on a previous term.

More generally, we can assume that $p(C)$ can be written as a product of probability terms $q_{t_i \in C}$, whether conditioned on previous terms or not. That is,

$$p(C) = q_{t_1 \in C} q_{t_2 \in C} q_{t_3 \in C} \cdots q_{t_N \in C} = \prod_{i=1}^{N} q_{t_i \in C}, \quad \text{Eq. (3)}$$

where $q_{t_i \in C}$ denotes the probability of occurrence of the token $t_i$ in claim $C$. A token $t_i$ is any convenient linguistic unit, which can for instance be a character, a word, or any textual element. The tokens considered depend on the language model chosen. For example, the tokens used by the GPT-$n$ networks for text generation include words, subwords (i.e., word components), and punctuation signs. In the present context, the tokens used to formulate probabilities associated with the claims can, in principle, be any element of writing.

### 3.2.2. Practical implementations with various language models

The following explores several ways of formulating the probability terms $q_{t_i \in C}$. Three basic probability models are devised in the foillowing: $p_0$, $p_1$, and $p_2$. The model $p_0$ relies on equal token probabilities. It is used to derive two practical models, $p_{wc}$ and $p_{cc}$, which are respectively based on the word count ("*wc*") and the character count ("*cc*") of a claim. The model $p_1$ exploits average token frequencies and is used to formulate $p_{wf}$ and $p_{cf}$, which respectively involve average word frequencies ("*wf*") and average character frequencies ("*cf*"). The model $p_2$ involves probability terms extracted from an LLM and is used to formulate a model noted $p_{LLM}$.

#### *3.2.2.1. Model of equal token probabilities*

The simplest approach is to assume that each token has the same probability $q_0$ of being drawn, irrespective of the particular field of the invention and the context set by other tokens in that claim. This amounts to replacing each term $q_{t_i \in C}$ by the same value $q_0$ in Eq. (3), such that the probability $p_0(C)$ associated with claim $C$ reduces to ${q_0}^N$. The corresponding scope function, $S_0(C)$, can thus be written as

$$S_0(C) = -\frac{1}{\log({q_0}^N)} = -\frac{1}{N} \cdot \frac{1}{\log(q_0)} = \frac{1}{N} \cdot \frac{1}{\log(1/q_0)}, \quad \text{Eq. (4)}$$

where $1/q_0$ corresponds to the total number (call it $N_{max}$) of tokens that can potentially be drawn.

The approximation leading to Eq. (4) amounts to considering the terms $q_{t_i \in C}$ as independent variables in Eq. (3) and replacing each of the corresponding logarithmic terms by some average value. This is consistent with the law of large numbers, which tells that $\log\left(\prod_i q_{t_i}\right) = \sum_i \log\left(q_{t_i}\right)$ converges almost surely to the expected value $\mu$ of each terms $\log(q_{t_i})$ times the number $N$ of tokens involved as $N \to \infty$.

As a result, $S_0(C)$ is proportional to the reciprocal of $N$, the number of tokens involved in claim $C$. If tokens are chosen to correspond to words, then the probability $p_{wc}$ is equal to $(1/N_{w,max})^{N_w}$, where $N_w$ is the word count of claim $C$ and $N_{w,max}$ is the total number of words that can potentially be drawn. The corresponding scope function $S_{wc}$ is thus proportional to the reciprocal of $N_w$, subject to the constant factor $1/\log(N_{w,max})$.

The same approach can be applied to any type of token. For instance, $N$ can also be chosen to correspond to the number $N_c$ of characters contained in claim $C$, yielding the probability $p_{cc}(C) = (1/N_{c,max})^{N_c}$ and scope function $S_{cc}(C) = 1/\big(N_c \log(N_{c,max})\big)$.

As one understands, scope calculations based on the reciprocal of the word or character count can be regarded as particular cases of Eq. (4), itself a particular case of Eq. (2). Thus, the methods used in [7, 8, 9, 10, 11] can be considered as falling under the definition of the scope function defined by Eq. (2) and Eq. (3). However, the present approach provides a more general definition of the claim scope and can accommodate probabilities obtained from any language model.

Interestingly, the number $N_c$ of characters can be thought of as encompassing more complete information than the sole number $N_w$ of words, because characters also include punctuation signs. Conversely, the character count is agnostic to the word count. For this reason, it is worth comparing scope values calculated from the word and character counts, as done in the next section. Hybrid approaches may also be contemplated, which are based on both $N_c$ and $N_w$.

More generally, Eq. (4) may be applied to the count of any text element of the claim, whether characters, words, phrase parts (noon phrases, verb phrases, etc.), or even claim features. However, claim features are not precisely defined and, thus, difficult to determine automatically. Similarly, Eq. (4) could also be applied to the number of nodes and/or edges in a graph representation of the claim [6, 15, 16], in place of the word or character count.

The main advantage of Eq. (4) is that it is very simple to implement, irrespective of the types of tokens chosen. However, a downside is that Eq. (4) only tells the probability of having $N$ tokens, in any combination. I.e., it covers any gibberish having the same token count as the claim of interest. And because Eq. (4) ignores the semantics attached to the tokens, it cannot discriminate between any two claims having an identical token count, hence the need for more sophisticated approaches.

#### *3.2.2.2. Word and character frequencies*

Anoter possibility is to rely on average frequencies $f_i$ of the respective tokens. This amounts to replacing each term $q_{w_i \in C}$ with $f_i$ in Eq. (3). The probability $p_1(C)$ can be written

$$p_1(C) = f_1 f_2 \cdots f_N = \prod_{i=1}^{N} f_i, \quad \text{Eq. (5)}$$

leading to

$$S_1(C) = -\frac{1}{\log(\prod_i f_i)} = -\frac{1}{\sum_i \log(f_i)}. \quad \text{Eq. (6)}$$

Use can be made of word frequencies, whereby each term $q_{w_i \in C}$ is replaced by a respective word frequency $f_{w_i}$. Such frequencies can be computed for large text corpora and can even be estimated for specified dates. E.g., the frequency of *smartphone* in 2023 is not the same as in 1997. If necessary, word frequencies can easily be estimated for specific technical fields, thanks to the literature available in these fields.

Using word frequencies, the probability $p_{wf}(C)$ of occurrence of a claim made of $N_w$ words can be written

$$p_{wf}(C) = f_{w_1} f_{w_2} \cdots f_{w_N} = \prod_{i=1}^{N_w} f_{w_i}. \quad \text{Eq. (7)}$$

The above probability ignores punctuation signs and is independent of the order of the words in the sentence. Thus, Eq. (7) reflects the probability of having any combination of the words $w_1, \cdots, w_N$, in any order. As such, it covers sentences that are syntactically incorrect, albeit composed of the same words as in the claim of interest, unlike Eq. (4).

The same approach can be applied to character frequencies $f_{c_i}$, which can easily be determined from text samples. Using frequencies of the $N_c$ characters involved in claim $C$, the corresponding probability $p_{cf}(C)$ can be written

$$p_{cf}(C) = f_{c_1} f_{c_2} \cdots f_{c_N} = \prod_{i=1}^{N_c} f_{c_i}. \quad \text{Eq. (8)}$$

This probability takes into account spaces and punctuation signs. It reflects the probability of having any combination of the same characters as involved in the claim of interest, in any order.

Eqs. (7) and (8) yield respective scope functions, namely

$$S_{wf}(C) = -\frac{1}{\sum_i \log(f_{w_i})} \text{ , and} \qquad \text{Eq. (9)}$$

$$S_{cf}(C) = -\frac{1}{\sum_i \log(f_{c_i})}. \qquad \text{Eq. (10)}$$

The above scope functions are based on average token frequencies. The $p_{cf}$ model cannot generate sensical words, while the $p_{wf}$ model ignores the part of speech variants and the context set by the combination of words. I.e., the $p_{wf}$ model is agnostic to the particular order and meanings of the words. In addition, the probability of each word in a sentence should be conditioned on the presence of other words in the sentence. For example, the probability of reading *the* after *of* (as in *of the*) is usually larger than the probability of reading *a* after *of* (as in *of a*).

#### *3.2.2.3. Large language models and conditional probabilities*

LLMs allow probability terms associated with each token to be more accurately evaluated in the context set by other tokens. For example, the GPT-2 completion network makes it possible to estimate token probabilities in the context set by previous tokens in a prompt [22]. That is, the probability of a token $t_i$ can be estimated in the context defined by all preceding tokens $\{t_1, \cdots, t_{i-1}\}$. So, the probability associated with claim $C$ can be written as a chain rule. I.e., each term $q_{t_i \in C}$ can be written as the conditional probability $q_{t_i|\{t_1,\cdots,t_{i-1}\}}$ of observing $t_i$ right after the ordered sequence $\{t_1, \cdots, t_{i-1}\}$, i.e., the “history” of the claim until token $t_i$. As a result, the overall probability $p_{LLM}(C)$ can be formulated as

$$p_{LLM}(C) = q_{t_1} q_{t_2|\{t_1\}} q_{t_3|\{t_1,t_2\}} \cdots q_{t_N|\{t_1,\cdots,t_{N-1}\}} = \prod_{i=1}^{N} q_{t_i|\{t_1,\cdots,t_{i-1}\}}. \qquad \text{Eq. (11)}$$

The first term $q_{t_1}$ can for instance be obtained from the frequency $f_1$ of the first word $w_1$ appearing in claim $C$, while the following terms are extracted from the LLM, see Section 3.4.1. Eq. (11) is closely related to the probability distribution of a word sequence in auto-regressive language generation tasks, where this probability decomposes into the product of conditional next word distributions [23]. However, here the task is not to generate language but evaluate, *a posteriori*, the probability associated with a certain text.

Plugging Eq. (11) into Eq. (2) yields

$$S_{LLM}(C) = -\frac{1}{\sum_i \log(q_{t_i|\{t_1,\cdots,t_{i-1}\}})}. \qquad \text{Eq. (12)}$$

As formulated above, the scope $S_{LLM}(C)$ takes into account the semantics attached to the words of the claim, but in a forward fashion only. I.e., the probability $q_{t_i|\{t_1,\cdots,t_{i-1}\}}$ of each token $t_i$ is only impacted by the sequence formed by the previous tokens $\{t_1,\cdots,t_{i-1}\}$.

For well-formed text, one expects an LLM to yield larger values for the probability terms $q_{t_i\in C}$ than the models based on averaged frequencies. This, in turn, should reduce the values of self-information and, thus, increase the scope values.

### 3.2.3. Properties of the scope function

The models described above assume that the probability associated with a claim $C$ can be written as the product of $N$ terms $q_{t_i\in C}$. Now, such a product can also be rewritten as the geometric mean of the terms $q_{t_i\in C}$ to the power of $N$, because the geometric mean $\langle q_{t_i\in C}\rangle_{\text{GM}} = \langle q_{t_1\in C}, q_{t_2\in C}, \cdots q_{t_N\in C}\rangle_{\text{GM}}$ is, by definition, equal to $\left(\prod_{i=1}^{N} q_{t_N\in C}\right)^{1/N}$. Thus, the scope $S(C)$ can be rewritten

$$S(C) = -1/\log\left(\prod_i q_{t_i\in C}\right) = -1/\left(N\log(\langle q_{t_i\in C}\rangle_{\text{GM}})\right) = \frac{1}{N}\cdot\frac{1}{\log\left(1/\langle q_{t_i\in C}\rangle_{\text{GM}}\right)} \qquad \text{Eq. (13)}$$

The function $S(C)$ is still proportional to the reciprocal of $N$, the number of tokens, albeit modulated by $1/\log\left(1/\langle q_{t_i\in C}\rangle_{\text{GM}}\right)$. This last term is constant where equal token probabilities are considered, i.e., $\langle q_{t_i\in C}\rangle_{\text{GM}} = q_0$ in Eq. (4). On the contrary, the term $1/\log\left(1/\langle q_{t_i\in C}\rangle_{\text{GM}}\right)$ depends on actual tokens when using average frequencies or LLM-based conditional probabilities. Although the average $\langle q_{t_i\in C}\rangle_{\text{GM}}$ can be substantially impacted by very small probability values, this dependence is mitigated by the logarithm function in the term $1/\log\left(1/\langle q_{t_i\in C}\rangle_{\text{GM}}\right)$, which varies slowly.

Thus, the scope values computed from Eq. (13) still depend primarily on the reciprocal of the token count, even where average token frequencies or LLM probabilities are used. However, advanced language models will be able to discriminate between claims having the same word or character count, as the value of $1/\log\left(1/\langle q_{t_i\in C}\rangle_{\text{GM}}\right)$ depends indirectly on the semantics attached to the text elements. That is, the semantics of the text elements impact their probabilities, which impact the scope of the claim.

### 3.3. Reformulation of the probability: disclaimers and alternatives

In practice, the single product in Eq. (3) is not best suited for addressing disclaimers and alternatives.

A disclaimer is a "negative" technical feature, which excludes a specific case from a more general feature. For example, a patent claim may be directed to an *electrical conductor made of any metal except gold*, where emphasis is put on the disclaimed subject-matter. While a language model trained on patent claim language would ideally interpret this exclusion, actual language models are unlikely to do so; they simply predict the probability of the next token [24].

Therefore, one may want to calculate the corresponding probability as the difference of two probability products, each defined according to Eq. (3). In the above example, the minuend and subtrahend respectively correspond to probabilities obtained for an *electrical conductor made of metal* and an *electrical conductor made of gold*.

Of course, this difference must be positive, something that is not guaranteed in practice. An LLM may possibly lead to a positive difference, thanks to the context provided (an *electrical conductor*), while simpler language models will likely fail. In the above example, a model based on word frequencies will lead to a negative difference because *gold* happens to be more frequent than *metal* in general literature, unlike patent literature. A model based on word counts will fail if the word count of the subtrahend is larger than or equal to that of the minuend, as in the above example. Similar remarks apply to character counts.

Similarly, the formulation of Eq. (3) does not lend itself well to alternatives. For example, a patent claim may recite an *electrical conductor made of copper or gold*, where emphasis is put on the two branches of the alternative. Such a formulation involves two alternative inventions, i.e., *an electrical conductor made of copper* and *an electrical conductor made of gold*. More generally, an alternative involves $n$ branches, which respectively correspond to $n$ alternative inventions. The scope of the resulting claim may be calculated independently for each branch. Alternatively, it may be calculated as the minimum, maximum, or average, of the probability products corresponding to the $n$ branches. A further possibility is to calculate the total scope of the claim, i.e., based on the sum of the probability products corresponding to the $n$ branches, as done in the following.

Thus, alternatives give rise to sums, whereas disclaimers give rise to differences of probability products. As one understands, the probability associated with a claim may be defined as a weighted sum of products of probability terms, where the weights can be positive or negative, to respectively account for alternatives and disclaimers. Thus, as opposed to the single product of Eq. (3), the probability $p(C)$ can more generally be written as a weighted sum of products of probability terms $q_{t_i \in C}$, involving $M$ terms, i.e.,

$$p(C) = \sum_{k=1}^{M} \omega_{k \in C} \prod_{i=1}^{N} q_{t_i \in k}, \qquad \text{Eq. (14)}$$

where $\omega_{k \in C}$ weights each branch in accordance with the occurrence of disclaimers and alternatives. Each disclaimer yields two branches, one of which is negatively weighted, whereas each alternative yields at least two branches, each being positively weighted. The resulting scope function writes

$$S(C) = -\frac{1}{\log\left(\sum_{k=1}^{M} \omega_{k \in C} \prod_{i=1}^{N} q_{t_i \in k}\right)}. \qquad \text{Eq. (15)}$$

Straightforward weighting schemes are used in this work. A same weight (equal to 1) is used for each branch of an alternative, assuming that the subject-matter of the branches are fully disjoint. In the case of a disclaimer, the minuend $m$ and the subtrahend $s$ are respectively weighted by 1 and –1, assuming that the subject-matter corresponding to the subtrahend is fully encompassed by that of the minuend. That is, $\omega_m = 1$ and $\omega_s = -1$.

One problem, however, is that there is no guarantee that the probability associated with $\omega_m$ is effectively larger than the probability associated with $\omega_s$, even when using an LLM. Alternative approaches can be contemplated. For instance, one may rely on ad hoc weighting schemes, whereby, e.g., $\omega_m$ may be set to some arbitrary number between 0.0 and 1.0, while $\omega_s$ is set equal to zero. NLP can be used to automatically identify branches in the claim. For example, string pattern recognition can be used to detect terminology typical of disclaimers or alternatives. Alternatively, this can also be done manually.

The scope function defined by Eq. (15) can be regarded as a generalization of Eq. (2) and Eq. (3). The other way around, Eq. (15) results from applying Eq. (3) to each branch of the claim and suitably weighing the corresponding contributions. Examples of calculations involving disclaimers and alternatives are discussed Section 4.1; they are based on Eq. (14) and Eq. (15) and use the straightforward weighting schemes discussed above. All other scope values are directly obtained from Eq. (2) and Eq. (3).

## 3.4. Calculation procedures and normalization

All computations are performed using the Wolfram Language and Mathematica 12.0 [25]. Natural logarithms (of base $e \approx 2.718$) are used in this work, whereby the unit of self-information is the nat and the scope is expressed in nat$^{-1}$. Choosing a different base merely amounts to rescaling the results.

The text of the claims is subject to minimal normalization. All claims are lowercased, before processing, except for acronyms (such as *DNA*) and other terms conventionally written in capital letters, if any. Any indefinite article at the beginning of a claim is deleted, for the sake of standardization and each claim starts with a space character. E.g., "*A writing instrument …*" and "*Writing instrument …*" are each transformed into "␣*writing instrument* … ", where "␣" denotes the space character. For completeness, the full stop of the sentence is discarded.

Seven probability models are considered. The probability of occurrence of each claim is computed as follows.

### 3.4.1. Large language models: GPT2, davinci-002, and Babbage-002

The first probability model ("*LLM*") is mostly used in conjunction with the GPT-2 transformer network [22]; the resulting model is referred to as the "GPT2" model hereafter. The GPT-*n* networks came into light with the boom of ChatGPT, which uses such networks to generate language in response to user prompts, according to an iterative process [26].

The GPT-2 network actually includes a family of individual nets. For the present work, this network is parameterized to extract probabilities, as described in [26], to allow probabilities of successive elements (words and punctuation signs) to be extracted in each claim. This way, the probability of occurrence of a claim can be formulated as a product, or a weighted sum of products, of probabilities of the successive elements, in accordance with Eq. (3) or Eq. (14).

To that aim, each claim is decomposed into elements including punctuation signs and words, and the probability of each element is recursively queried in the context defined by the previous elements. In detail, an average word frequency is used to estimate the probability of the very first word. This word serves as context to determine the probability of the second element (a word or a punctuation sign). The GPT-2 network used herein can generate probability values for any element in its vocabulary, which makes it possible to identify the probability value associated with the second element. Next, the first and second element are used as new context to determine the probability of the third element, and so on. This, eventually, makes it possible to formulate the global probability of the claim as a product, or a sum of products, of probability values obtained for each element in the claim.

A cutoff probability is used for those rare words for which no probability can be identified. That is, where no probability can be identified for a particular word, the probability of this word is taken to be equal to the lowest value of the probability terms identified for other elements in the claim.

The GPT2 results reported herein were obtained with a small version of the GPT-2 network, which involves 117 million parameters only [27]. Tests carried out with the medium-sized version (355 million parameters) did not show significant changes over the small version, it being reminded that the goal here is to compute log-probabilities and not to generate language. More recent OpenAI models like GPT-3.5 and GPT-4 are not open, which prohibits a direct computation of the token probabilities.

While a variety of LLMs are available, which often allow log-probabilities of the generated responses to be extracted, few models allow log-probabilities to be extracted from prompts, as needed in the present context. In particular, the OpenAI API [28] no longer permits extraction of log-probabilities associated with tokens in the prompts, except for two pure completion models, namely the so-called babbage-002 model (1.3 billion parameters) and davinci-002 model (175 billion parameters), which are replacements for the original GPT-3 base models [29].

Thus, the GPT2 calculations were complemented by results obtained through the OpenAI API based on the Babbage-002 and Davinci-002 models. As with the GPT2 results, an average word frequency is used to estimate the probability of the very first word, while subsequent token log-probabilities are obtained through the OpenAI API, by sending a “completion” request to the OpenAI API, using parameters “Echo” and “LogProbs” respectively set to “True” and “1”. However, the OpenAI API did not make it possible to replicate the procedure used to control the cutoff probability for very rare words as discussed above with respect to the GPT-2 network.

### 3.4.2. Word frequencies

The second probability model (“*wf*”) is based on word frequencies; spaces and punctuation signs are ignored. The probability associated with each claim is formulated as a product, or a sum of products, of the average frequencies of the words involved in the claim, see Eq. (7). Again, a cutoff probability is used for words for which no frequency is known. In the present work, this probability was set to $7.59559\ 10^{-11}$, which corresponds the minimal frequency for the used set of available word frequencies [25].

### 3.4.3. Word count

The third probability model ("*wc*") is based on the word count $N_w$ of (each branch of) the claim, in accordance with Eq. (4), where $N$ is set equal to $N_w$. In Eq. (4), the term $1/q_0$ represents a total number $N_{w,max}$ of words that may be drawn at random from a uniform distribution. In the present context, this number should ideally represent the total number of words in the English technical language, which may be used in a patent claim. This number was set equal to 155327, in accordance with the number of words contained in WordNet database [30].

The scope is calculated as the reciprocal of the number $N_w$ of words, times a multiplicative factor equal to $1/\text{Log}(N_{w,max})$. Note, the exact number $N_{w,max}$ considered is not critical as it is downplayed by the logarithm. Also, this scaling factor $N_{w,max}$ only impacts the magnitudes of the scope values obtained, globally, and not their ratios.

### 3.4.4. Character frequencies

The fourth probability model is based on character frequencies ("*cf*"). The probability associated with each claim is formulated as a product, or a weighted sum of products, of average frequencies of the characters involved in the claim. All claim characters are considered, including punctuation signs. The frequencies considered were obtained from a compilation of Wikipedia articles concerning patents, mechanics, physics, chemistry, biology, computer science, and mathematics [31].

### 3.4.5. Character count

The fifth probability model ("*cc*") is based on the character count $N_c$ of (each branch of) the claim. I.e., the scope is calculated as the reciprocal of the number $N_c$ of characters, times a multiplicative factor. The latter is equal to $1/\text{Log}(N_{c,max})$, where $N_{c,max}$ is an estimate of the total number of characters commonly used in claims in the English language. In the present work, this number is fairly arbitrarily taken to be equal to 50, corresponding to {" ", ",", ".", ":", ";", "-", "+", "–", "×", "%", "(", ")", "0", "1", "2", "3", "4", "5", "6", "7", "8", "9", "/", "%", "a", "b", "c", "d", "e", "f", "g", "h", "i", "j", "k", "l", "m", "n", "o", "p", "q", "r", "s", "t", "u", "v", "w", "x", "y", "z"}. Again, the actual number chosen for $N_{c,max}$ is downplayed by the logarithm and only impacts the scope values globally, without affecting ratios.

## 4. Applications to actual patent claims

Scope values are now calculated for multiple sets of patent claims, using the various probability models discussed in Section 3.

### 4.1. Application to a set of simple claims

Scope values are first calculated for a set of short patent claims, using six of the probability models described above, i.e., the models based on davinci-002, GPT-2, word frequencies, word counts, character frequencies, and character counts. The results from the babbage-002 model are not reported because they are essentially redundant with the davinci-002 results. The claims were devised to synthetically illustrate the strengths and weaknesses of the present approach. The results are presented in Table 1.

| Claims | Calculated scope values ($\times 10^5$) | | | | | |
|---|---|---|---|---|---|---|
| | Davinci-002 | GPT2 | Word Freq. | Word count | Char. Freq. | Char. Count |
| Claim of gradually decreasing scope | | | | | | |
| *1. Hammer including a handle and a metal head.* | 2556 | 2364 | 1628 | 1046 | 804 | 594 |
| *2. Hammer including a handle and a steel head.* | 2440 | 2306 | 1629 | 1046 | 812 | 594 |
| *3. Hammer including a handle and a stainless steel head.* | 2310 | 2232 | 1360 | 930 | 662 | 482 |
| Claim with an alternative | | | | | | |
| *4. Hammer including a handle and a head of steel or aluminum.* | 1946 | 1888 | 1226 | 761 | 593 | 441 |
| *4a. Hammer including a handle and a head of steel or aluminum.** | 2443 | 2357 | 1630 | 1053 | 812 | 594 |
| Claim with a disclaimer | | | | | | |
| *5. Hammer including a handle and a head of metal except steel.* | 1726 | 1617 | 1199 | 761 | 585 | 433 |
| *5a. Hammer including a handle and a head of metal except steel.** | 2545 | 2340 | - | - | - | - |
| Patent terminologies: "*including*" vs. "*comprising*" vs. "*consisting of*" | | | | | | |
| *6. Hammer including a handle and a metal head.* | 2556 | 2364 | 1628 | 1046 | 804 | 594 |
| *7. Hammer comprising a handle and a metal head.* | 2634 | 2112 | 1565 | 1046 | 787 | 581 |
| *8. Hammer consisting of a handle and a metal head.* | 2791 | 2180 | 1495 | 930 | 750 | 544 |

Table 1. Calculated claim scope values of short claims computed from various probability models (based on davinci-002, GPT-2, word frequencies, word counts, character frequencies, and character counts) defined according to Eq. (2), except (*) for claims 4a and 5a, where use is made of Eq. (14).

The claims are directed to a *hammer*, which basically includes a *handle* and a *metal head* (claim 1). Claims 2 and 3 gradually refine the technical features introduced in claim 1: claim 2 recites a *steel head* and claim 3 narrows down the steel head to a *stainless steel_head*. These successive scope restrictions are consistently reflected by the Davinci-002 and GPT2 models. The other models fail to show a loss of scope from claim 1 to claim 2. The reasons are that such claims contain the same number of words and characters, and the word *steel* happens to be more frequent than the word *metal* in literature.

The next two claims (4 and 4a) include an alternative: the head is made of *steel or aluminum*. Claims 4 and 4a are identical, but subject to different probability calculations. In the first case (claim 4), the probability is defined as a single product, Eq. (2), as with claims 1 to 3. There, all probability models, including davinci-002 and GPT2, fail to predict that the alternative of claim 4 (*steel or aluminum*) is necessarily broader than claim 2, reciting a *steel head*. This is mainly due to the relative increase of the word count from claim 2 to claim 4, which reduces the probabilities for claim 4.

As noted in Section 3.3, alternatives can more suitably be addressed by defining the probability as a sum of products, Eq. (14). Doing so resurrects the expected scope hierarchy, at least when using LLMs (davinci-002 and GPT2) and the word-based models; compare the scope values listed in respect of claims 2 and 4a. In addition, comparing the results for claims 1 and 4a shows that the LLMs correctly predict that a *metal head* is broader than a *head of steel or aluminum*. The other models fail due to unfavorable ratios of counts or word frequencies.

Claims 5 and 5a are identical, too, and include a disclaimer: the head is made of *metal except steel*. In the first case (claim 5), the probability is taken as a single product, as with claims 1 to 4. This exaggeratedly reduces the calculated probabilities and, in turn, the scope value. When using instead a difference of probability products (see the results for claim 5a), the LLMs give rise to scope values that are indeed smaller than that of claim 1. All the other models fail, due to a difference of probabilities that is negative or equal to zero.

Claims 6, 7, and 8, compare results obtained with the terminologies *including*, *comprising*, and *consisting of* (claim 6 is identical to claim 1). In patent law, *including* and *comprising* are synonyms: they mean that all the mentioned features are encompassed, without precluding the potential presence of additional unspecified features. On the contrary, *consisting of* restricts to those features that are explicitly specified in the claim [3]. Thus, the scope of claim 6 or 7 is, in principle, larger than that of claim 8, something that LLMs fail to reflect consistently. One reason might be that the wording *consisting of* is more common than *comprising*, as queries in search engines confirm. The other models perform better here, which is a fortunate consequence of the favorable differences in terms of word or character counts.

Thus, the results shown in Table 1 synthetically capture the main outcomes of the present approach. Probabilities extracted from language models can be used to estimate scope values and the results obtained from LLMs are globally more coherent than those obtained from simpler language models. However, pretrained LLMs will occasionally fail as they were not specifically trained on patent claim language. Similarly, average word frequencies extracted from general corpora are not systematically higher for broader terms, even though they mostly are, as discussed in Section 4.5.

Alternatives and disclaimers will typically be an issue when calculating the scope based on Eq. (2) and (3). The variant captured by Eqs. (14) and (15) benefits the results, to some extent. However, further work appears to be necessary in this area. In particular, ad hoc schemes may be used to suitably normalize the weights associated with the branches of the claims to ensure consistent results, as discussed in Section 3.4. Alternatives and disclaimers are no longer considered in the following, it being noted that such cases only concern a minority of independent claims in practice.

### 4.2. Calculated scope of various series of gradually narrowed claims

A systematic investigation is in order. Nine series of claims are considered, which relate to various technical fields, i.e., mechanics, chemistry, biochemistry, material sciences, electricity, and computer sciences. Each series contains between 7 and 22 claims, the scope of which is gradually narrowed down by successively adding or refining claim features of the preceding claims. As a result, the claims of each series mostly have increasing word and character counts. Overall, the claims contain between 1 and 320 words. They were adapted from real patents or patent applications, or fictive examples of inventions as proposed in the framework of the European Qualifying Examination [14].

The nine series of claims are listed in Sections A.1 to A.9 of Appendix A. The corresponding scope values are reported in Appendix B. FIGS. 1 and 2.1 show values of scope and self-information computed from the GPT2 model for each claim of the nine series. Such values are depicted against the word count, i.e., the number of words in each of the claims considered. The plot ranges are truncated on the *x*-axis, for the sake of depiction; see Appendix B for the complete set of calculated scope values.

By construction, the values shown in FIG. 1 are the reciprocal of the self-information values shown in FIG. 2.1. In general, the scope values (FIG. 1) decrease as the amount of information increases, which mostly translates into an increasing word count. I.e., the more information contained in a claim, the smaller the corresponding scope value. Interestingly, FIGS. 1 and 2.1 also show that, for a given word count value, claims directed to different inventions result in different scope values and different values of self-information. That is, for a value on the *x*-axis (e.g., 40 words), claims in distinct series give rise to different amounts of self-information and, in turn, different scope values. This results from the fact that the GPT2 model takes the semantics into account, if only indirectly. Similar observations can be made when depicting the curves against the character counts (not shown).

On the contrary, self-information values obtained from a simple model based on the reciprocal of the word count are necessarily identical for any fixed word count value, irrespective of the invention claimed. Such a simple model results in overlapping self-information curves (in fact straight lines, see FIG. 2.3), such that the scope values obtained for claims having the same word count are identical, irrespective of the inventions claimed.

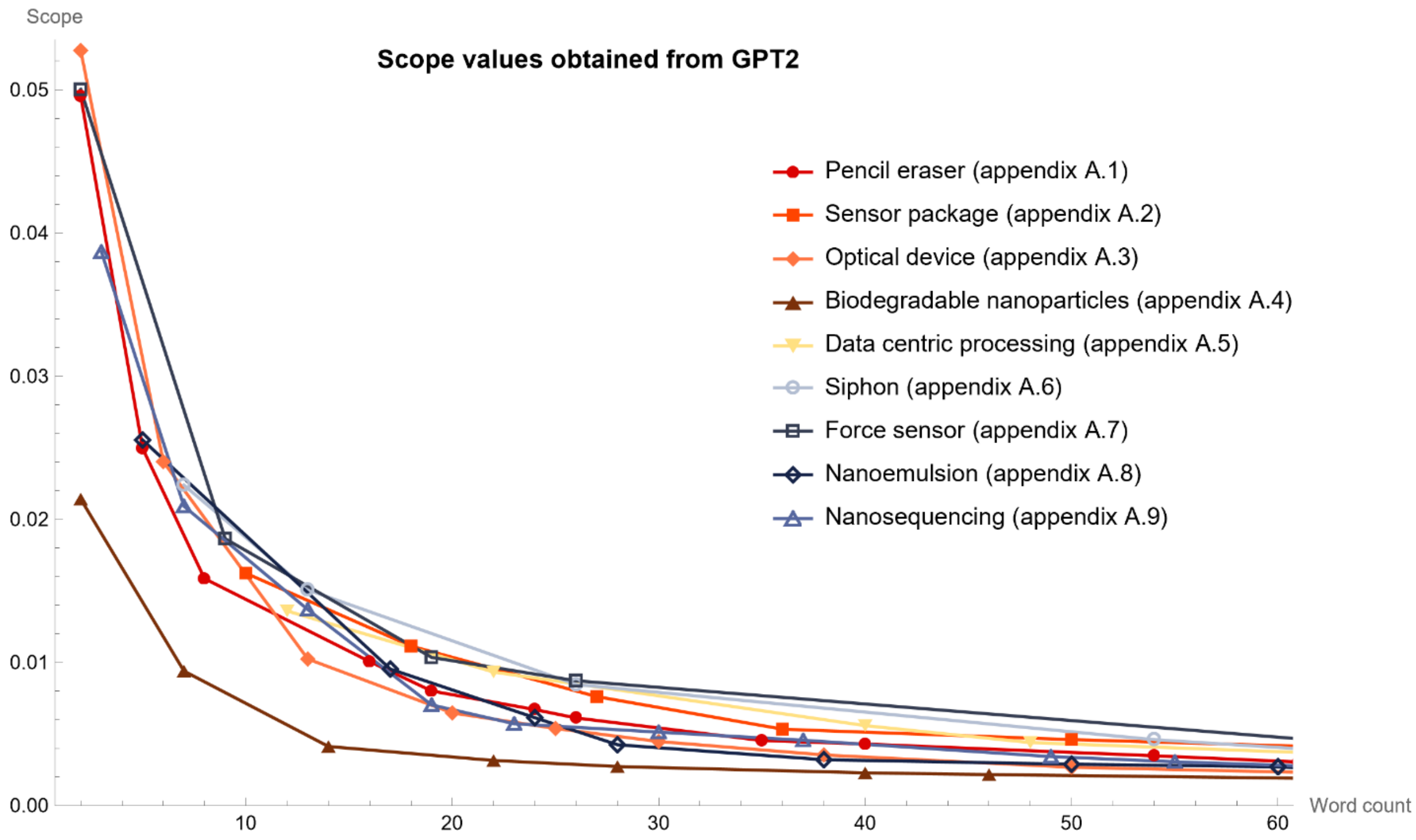

FIG. 1: Plot of claim scope values ($y$-axis) obtained from probabilities computed with the GPT2 network. The scope values ($y$-axis) are depicted as a function of the word count ($x$-axis) of the claims in each of the nine series of claims listed in Sections A.1 to A.9 of Appendix A.

Now, there is no reason for this to be true in practice: claims directed to different inventions would be expected to have different scopes, even if they have the same word count.

The other probability models result in curves that are more or less dispersed, though less than the GPT2 curves (compare FIG. 2.1 with FIGS. 2.2, 2.4, and 2.5). That is, such models can discriminate between claims having identical word counts, to some extent. The larger the dispersion, the better the propensity of the model to discriminate between claims directed to distinct inventions.

This may be surprising at first sight for the model based on the character count, see FIG. 2.5. However, one reason for this is that longer words (i.e., words having more characters) tend to be not only rarer but also conceptually more complex on average [32].

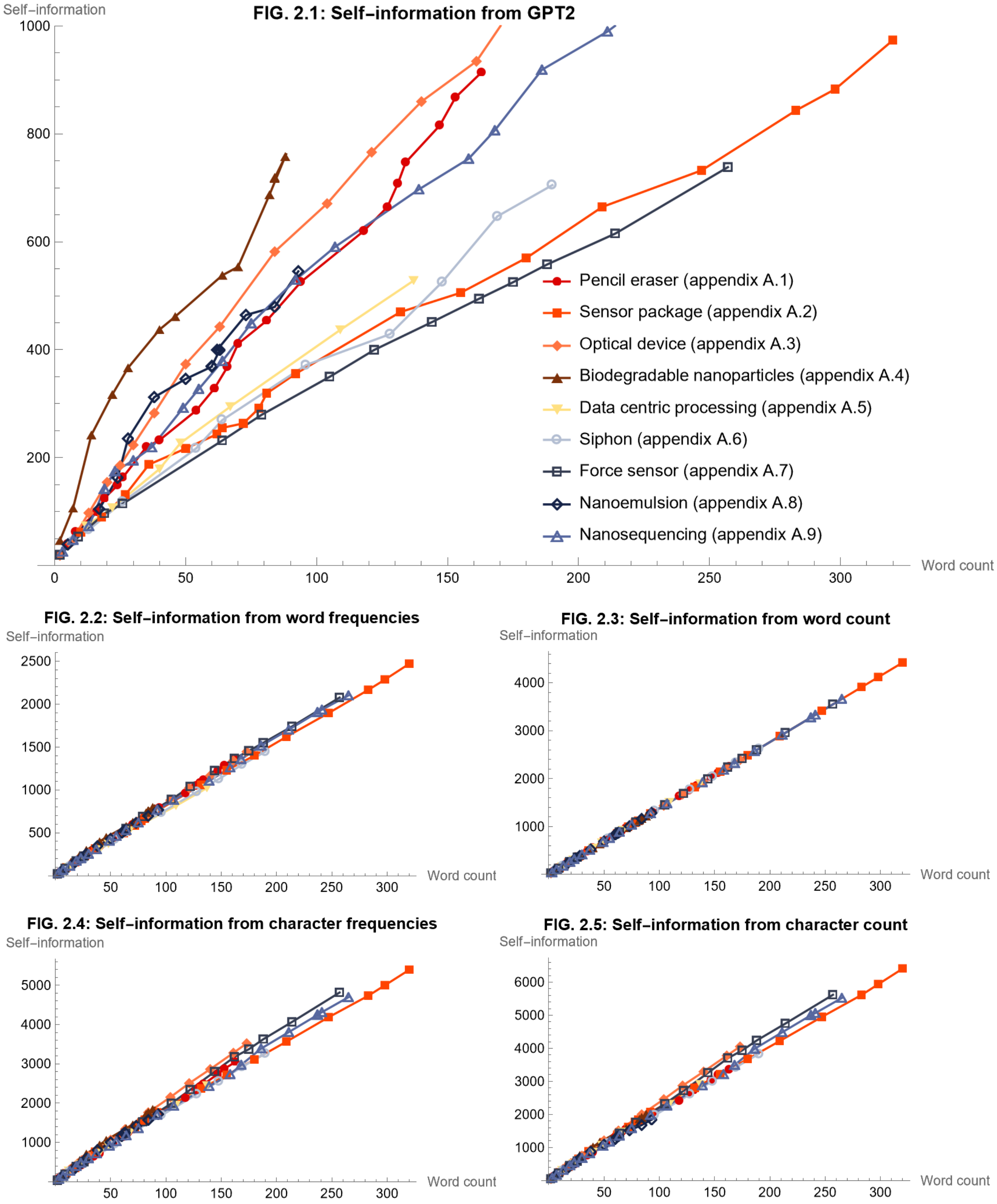

FIG. 2: Self-information values (y-axis) obtained from probability models based on various language models (FIG. 2.1: GPT2, FIG. 2.2: word frequencies, FIG. 2.3: word counts, FIG. 2.4: character frequencies, FIG. 2.5: character counts). FIG. 2.1 corresponds to the scope values of FIG. 1. The self-information values ($y$-axis) are depicted as a function of the word count ($x$-axis) of the claims in each of the nine series of claims listed in Sections A.1 to A.9 of Appendix A.

FIG. 3 shows word frequencies of randomly selected words as a function of the corresponding character counts. Up to 400 words are randomly selected for each character count value ($x$-axis). In total, 4946 words were sampled, which corresponds to approximately 13% of the total number of words available in the built-in dictionary used [25]. The orange squares denote the mean word frequency values for each character count value, illustrating that longer words tend to be rarer.

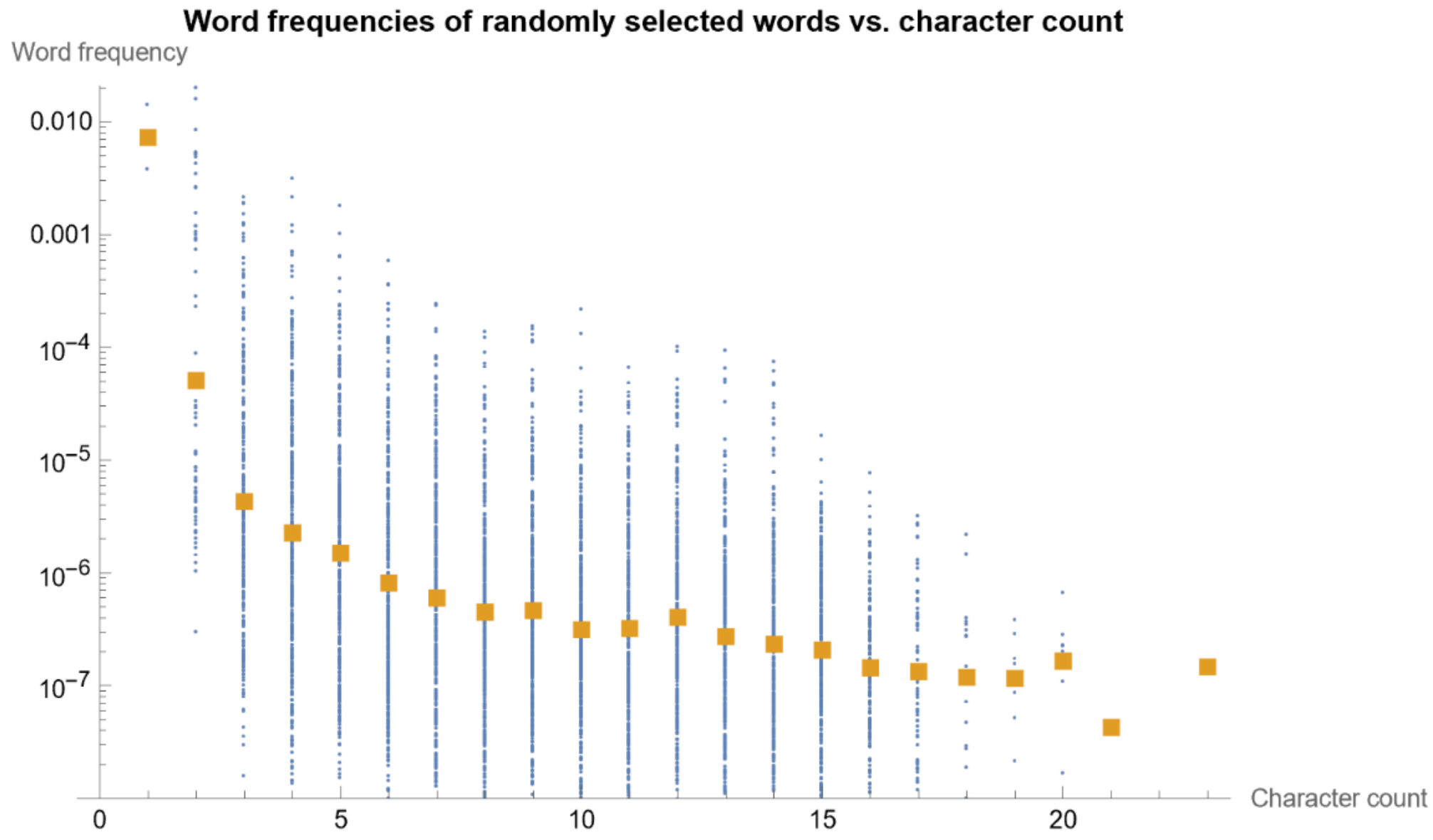

FIG. 3: Word frequencies (log scale) of randomly selected words (blue dots) as a function of the corresponding character counts. The orange squares denote the mean values for each character count value.

A model based on character counts cannot discriminate between any two claims having an identical character count. However, the probability that two claims have the same character count is less than the probability that they have the same word count. For instance, tests performed on random samples of 10,000 independent claims to distinct inventions show that less than 1% of the patent claims have the same word count, while the number of collisions in terms of characters is one order of magnitude less.

Next, FIG. 4 compares score values obtained from the seven probability models, for two series of claims directed to a sensor package and a nanoemulsion, see Appendix A.2 and A.8. Similar patterns are observed for the other claim series (not shown). The $x$-axis now corresponds to the claim number in the series and can roughly be equated to the number of technical features involved in the corresponding claim. Again, a claim feature is not precisely defined. However, each series of claims was devised so that each claim in each series adds or refines a technical feature of the previous claim in this series. As a result, the scope of the claims gradually narrows down along the $x$-axis.

In general, the more sophisticated the probability model, the higher the scope values. This is consistent with the observation made in Section 3.2.2.3: a more advanced the language model results in higher probability value for well-formed text. Higher probabilities translate into lower self-information values and, in turn, higher scope values.

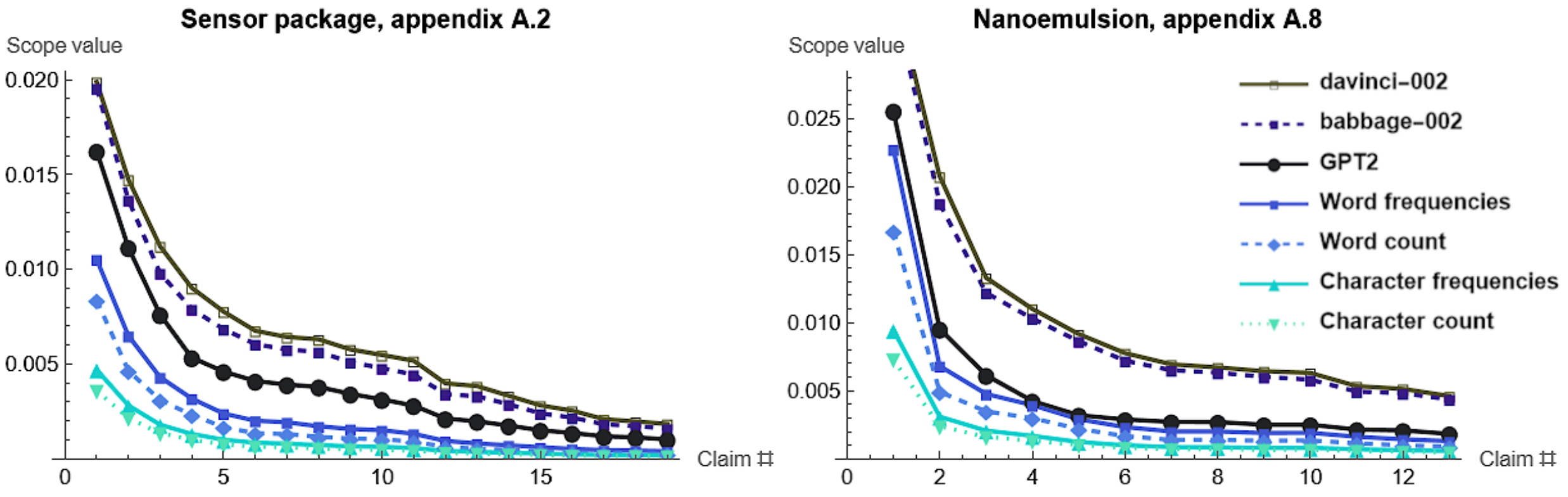

FIG. 4: Claim scope values ($y$-axis) obtained based on probabilities from the various probability models (davinci-002, babbage-002, GPT2, word frequencies, word counts, character frequencies, and character counts) for two series of claims, respectively directed to a sensor package and a nanoemulsion, see Appendix A.2 and A.8. The $x$-axis corresponds to the claim number in the respective series.

Accordingly, the scope values obtained from davinci-002 model are generally larger than the babbage-002 scope values, which are themselves larger than the GPT2 values. Similarly, the scope values obtained from the LLMs are generally larger than the scope values obtained from word frequencies, which are themselves larger than the values obtained from character frequencies. The results obtained from the models based on the word and character counts cannot be directly compared as their scope values are scaled, see Section 3.4.

| | davinci-002 | babbage-002 | GPT2 | Word Freq. | Word Count | Char. Freq. | Char. Count |
|---|---|---|---|---|---|---|---|
| davinci-002 | 1.000 | 0.998 | 0.969 | 0.965 | 0.947 | 0.949 | 0.955 |
| babbage-002 | | 1.000 | 0.972 | 0.969 | 0.949 | 0.952 | 0.959 |
| GPT2 | | | 1.000 | 0.971 | 0.937 | 0.971 | 0.974 |
| Word Freq. | | | | 1.000 | 0.990 | 0.977 | 0.983 |
| Word Count | | | | | 1.000 | 0.961 | 0.967 |
| Char. Freq. | | | | | | 1.000 | 0.999 |
| Char. Count | | | | | | | 1.000 |

Table 2. Pearson correlation coefficients between scope values obtained based on probabilities from various probability models (davinci-002, babbage-002, GPT2, word frequencies, word count, character frequencies, character count) for all claims of the nine series listed in Sections A.1 to A.9 of Appendix A.

All in all, the scope values obtained from the various probability models are highly correlated, see Table 2. Note, the scaling factors used in the models based on the word and character counts do not impact the correlation values. Compared with the word count values (second column), the scope values obtained from the character counts better correlate with the LLM results. That said, the correlation values are all very close to 1. One reason is that all probability models considered herein are primarily impacted by the numbers *N* of tokens involved, and the ratios of token counts used by the various models are roughly constant across the claim series.

### 4.3. Standard deviations for sets of claims of identical scope

A first experiment is performed, where multiple claim variations are generated for some of the inventions discussed in Section 4.2. Claim variations are obtained by combining alternative sentence parts (see Appendix C.2) to form tuples. As a simple example, the set {{"*Writing instrument*"}, {"*comprising*", ", *which comprises*"}, {"*a pencil*"}} can be used to generate 1 × 2 × 1 claim variations, namely *Writing instrument comprising a pencil* and *Writing instrument, which comprises a pencil*. The resulting claims variations involve the exact same technical features and thus have the exact same scope, even if they do not have the same word count.

| | | Reduced standard deviation of calculated scope values | | | | | | |
|---|---|---|---|---|---|---|---|---|
| Claim variations on | Average word count | davinci -002 | babbage -002 | GPT2 | Word Freq. | Word Count | Char. Freq. | Char. Count |
| Pencil-eraser | 23.5 | 3.84 | 4.92 | 2.56 | 7.18 | 7.13 | 8.28 | 8.56 |
| Optical device | 58.9 | 2.45 | 3.05 | 1.95 | 2.58 | 3.14 | 2.86 | 2.87 |
| Biodegradable nanoparticles | 41.2 | 4.55 | 4.84 | 2.80 | 3.42 | 5.66 | 4.13 | 4.4 |
| Datacentric computing | 68.4 | 1.86 | 1.70 | 3.24 | 2.83 | 2.88 | 2.97 | 2.91 |
| Nanoemulsion | 61.9 | 2.55 | 3.25 | 1.22 | 2.41 | 2.73 | 2.46 | 2.74 |
| Average | | 3.05 | 3.55 | 2.35 | 3.69 | 4.31 | 4.14 | 4.29 |

Table 3. Reduced standard deviation (%) of scope values obtained according to different probability models (based on davinci-002, babbage-002, GPT2, word frequencies, word counts, character frequencies, and character counts) for five sets of 96 claim variations on respective inventions.

Similarly, the lists of alternative sentence parts in Appendix C.2 are used to generate five sets of claims variations, directed to different inventions. Next, 96 claims variations are randomly sampled in each set. By construction, the claims sampled in any of the five sets are directed to the exact same invention and, thus, have the exact same scope, although they do not necessarily have the same word count. They should ideally result in the exact same scope value, such that their standard deviations should ideally be equal to zero. In practice, however, the calculated scope values differ from one claim to the other in each series. The differences observed depend on the probability model used, thereby illustrating limitations of such models. One way to assess this is to measure the reduced standard deviation (RSD) of the scope values in each claim set. Table 3 shows the RSD values (as percentages) obtained for each of the seven models discussed earlier.

In general, the LLMs result in lower reduced standard deviations, see Table 3. The GPT2 probability model outperforms the word frequency model, itself ahead of both character-based models. The model based on word counts comes last. However, the results obtained from davinci-002 and babbage-002 do not improve over GPT2, something that may be due to the use of different cutoffs, see section 3.4.1.

The results of Table 3 should be cautiously interpreted. That the RSD values are relatively small does not mean that the models can be considered acceptable for comparing the scope of any two specific claims. Such RSD values only measure average dispersions; the calculated scope values can substantially vary from one claim to the other, particularly when using a model based on the word or character count. For instance, while claims 1a and 1b in Section 2.2 have the exact same scope, the relative differences of scope values are equal to 17% (GPT2), 57% (word frequencies), 53% (word counts), 57% (character frequencies), and 56% (character counts).

### 4.4. Calculated scope values for a set of claims having an identical word count

Conversely, one may, for the same invention, devise claims having the same word count but a different scope. I.e., the claims recite more or less detailed features with different amounts of information, notwithstanding their constant word count. Preliminary experiments have been performed on a set of eight claims, each consisting of 25 words, directed to a pencil eraser, see Appendix C.1.

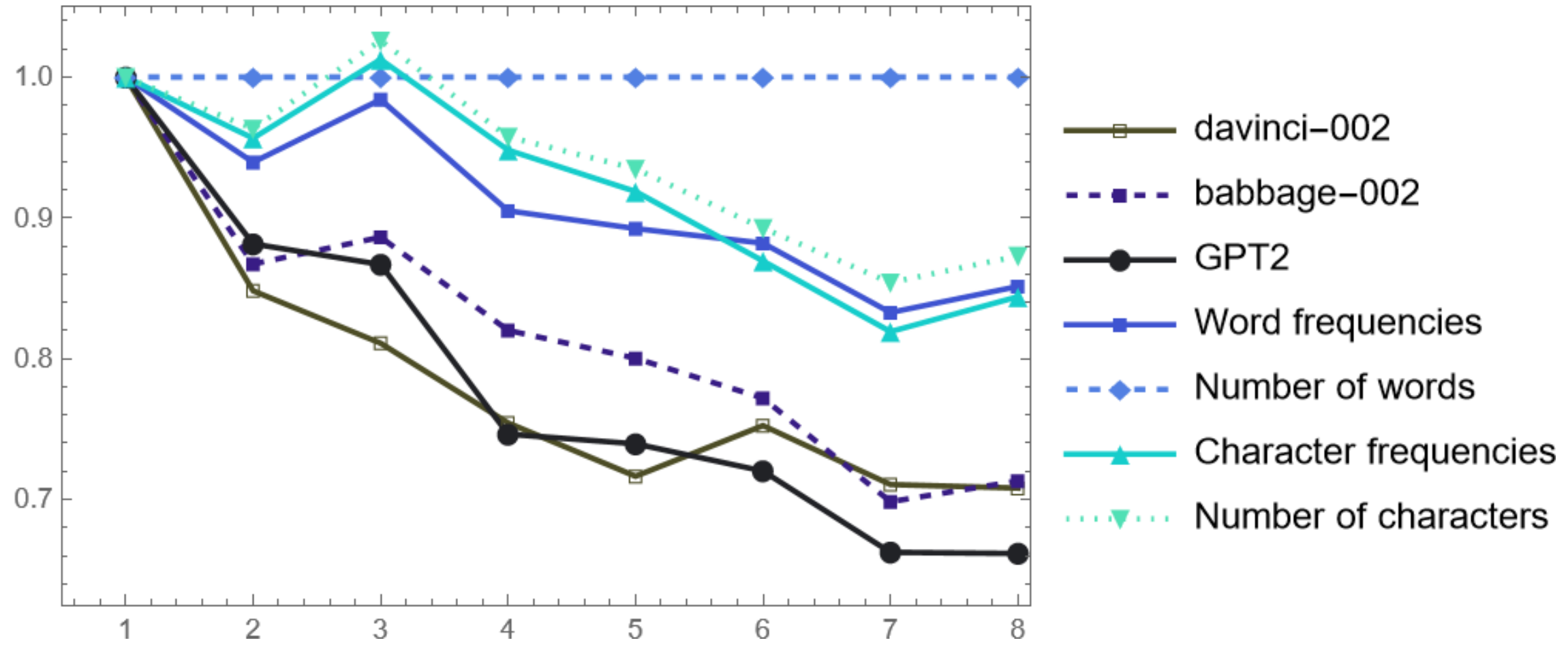

FIG. 5: Normalized scope values obtained for 8 variants of claims directed to a pencil eraser, and sorted from the broadest to the narrowest claim, using the same probability models as in FIG. 4.

The eight claims are sorted from the broadest to the narrowest claim. FIG. 5 depicts the scope values obtained from the seven probability models for the sorted claims. The values obtained are normalized to the scope value obtained for the broadest claim (the very first claim). The scope values obtained from the GPT2 probability model systematically decrease, unlike values obtained from the other models. Obviously, the model based on the word counts cannot discriminate between the claims (each consisting of 25 words), which results in a flat curve. Similarly, the model based on the character counts could not discriminate between claims of the same character count. However, the character counts of the claims mostly increase throughout the sorted claims in that case. As a result, this model can discriminate between the claims, like the models based on frequencies. The simpler models, however, do not correctly predict a systematic decrease of scope, unlike the GPT2 probability model. The davinci-002 and babbage-002 model also fail to systematically predict a decrease of the scope, something that could be due to the cutoff issue noted earlier.

### 4.5. Assessing inequalities between calculated scope values against known hierarchies of terms

Scope values are now calculated for known hierarchies of terms, thanks to a dictionary listing narrower terms of each entry. This dictionary is a built-in dictionary of Mathematica, which was compiled from a wide range of sources [33]. Once the hierarchies are built, the inequalities between the calculated scope values can be compared with the inequalities implied by the known hierarchical relations.

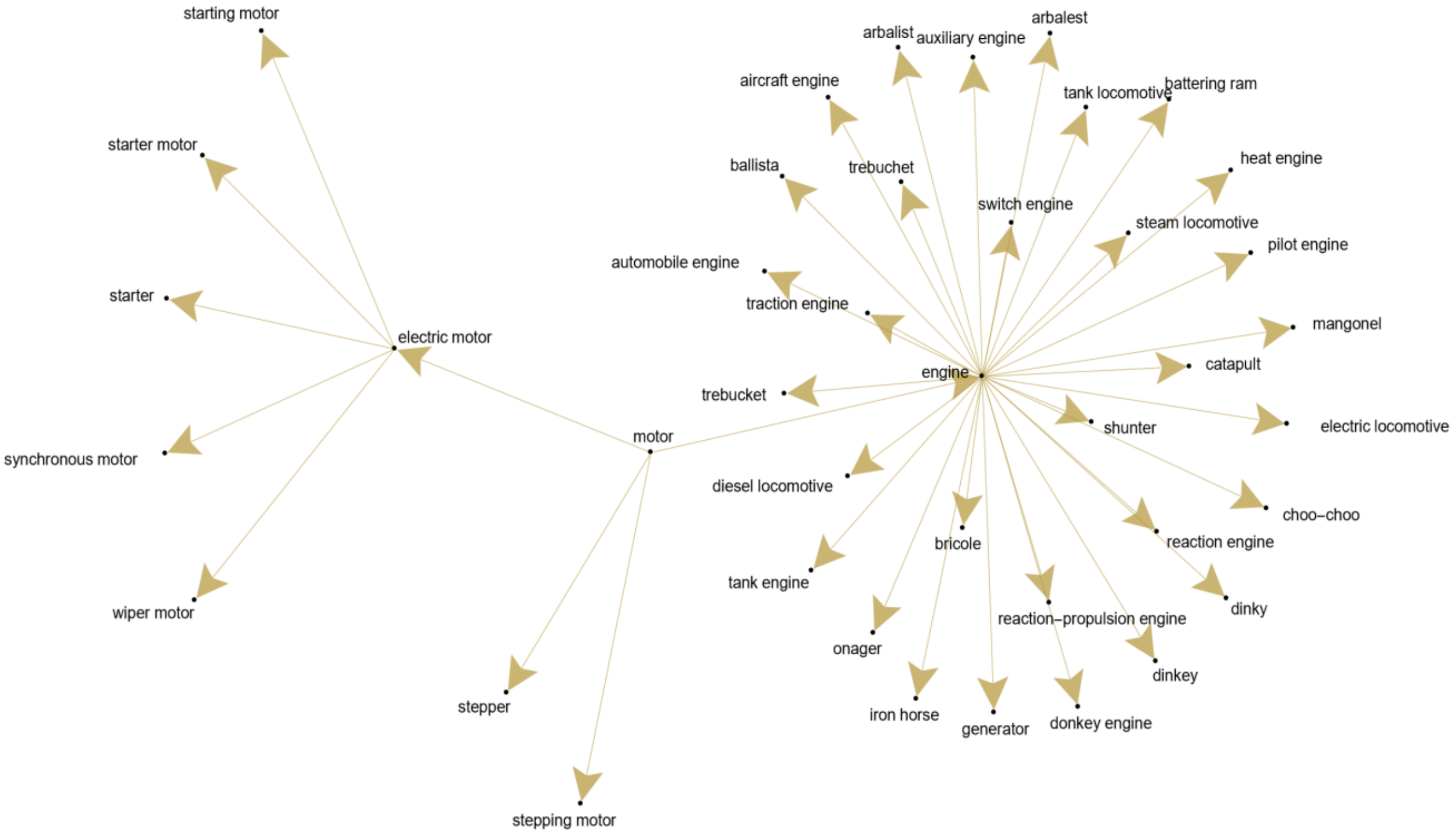

FIG. 6: Graph capturing hierarchical relations between terms ordered from a root representing the broadest term (*motor*), where each child node corresponds to a term having a narrower scope than its parent node.

Practically speaking, each hierarchy is constructed as a graph extending from a given root term, corresponding to the broadest term. Each child node corresponds to a term having a narrower scope than its parent node, by construction. Each pair of parent – child nodes forms hypernym – hyponym pair, which implies a hierarchical relation. An example is depicted in FIG. 6, where the root term used is *motor* and the depth of the graph is equal to two. As seen in the resulting graph, *motor* is broader than *electric motor*, which is itself broader than *synchronous motor*. Then, the relations between the calculated scope values are compared with the known hierarchical relations.

The results are compiled in Tables 3 to 5. Each table indicates the percentage of verified hierarchical relations, for each probability model. A known hierarchical relation between two given terms is deemed to be verified if it is consistent with the inequality obtained between the corresponding scope values. For example, the scope value of *electric motor* should be larger than that of *synchronous motor*.

The compared terms mostly consist of single words; they occasionally include two terms, more rarely three terms. Therefore, the results produced by any LLM are mostly identical to results obtained from the word frequency model in that case. For this reason, only the GPT2 results are reported below; results obtained with the davinci-002 and babbage-002 models are not reported as they are very similar, if not identical, to the GPT2 results.

In general, the scope values calculated with the probability models based on GPT2 and word frequencies agree very well with the known hierarchical relations (in more than 90% of the cases). There are unfortunate exceptions (e.g., *metal* vs. *steel*, as discussed in Section 4.1). The GPT2 model does not substantially improve on the word frequency model because most of the compared terms consist of single words, as noted above. The character-based models yield satisfactory results, too, while the model based on word counts fails in the majority of the cases. The reason is that most of the compared terms consist, each, of a single word, such that their scope values are identical. Still, in 43.8% of the cases, the word counts of the child node (e.g., 2 for *electric motor*) exceeds that of the parent node (e.g., 1 for *motor*), which leads to a fortunate match.

The fact that the root nodes mostly consist of single-word terms favorably impacts the results obtained, because all the probability models primarily influenced by the token count, as discussed in Section 3.2.3. Conversely, using circumlocutions (e.g., an *information processing system*), instead of more concise synonyms (e.g., *computer*), negatively impacts the results: the token counts of parent nodes corresponding to circumlocutions are usually larger than the word counts of the respective child nodes, as illustrated in Table 5. This explains the failures observed when using roots such as *computing device*, *information processing system*, or *hand tool*, even with the GPT2 model.

| Category | Root (tree depth) | Number of tested relations | Percentages of relations verified: GPT2 | Word frequency | Word count | Character frequency | Character count |
|---|---|---|---|---|---|---|---|
| Chemistry | Compound (1) | 110 | 94.7 | 94.8 | 15.5 | 49.1 | 52.6 |
| | Magnesium compounds (all listed compounds) | 224 | 100.0 | 100.0 | 97.3 | 99.6 | 99.6 |
| | Silicon compounds (all listed compounds) | 984 | 99.9 | 99.9 | 68.9 | 99.7 | 99.6 |
| Electronics | Circuit (1) | 35 | 100.0 | 100.0 | 60.0 | 85.7 | 80.0 |
| | Semiconductor (3) | 322 | 83.7 | 83.5 | 48.8 | 62.4 | 61.8 |
| | Transistor (1) | 3 | 100.0 | 100.0 | 66.7 | 66.7 | 66.7 |
| ICT | Computer (8) | 138 | 84.1 | 82.6 | 58.7 | 79.0 | 76.1 |
| | Information (1) | 64 | 100.0 | 100.0 | 20.3 | 34.4 | 26.6 |
| | Network (1) | 26 | 100.0 | 100.0 | 34.6 | 61.5 | 65.4 |
| | Phone (1) | 29 | 96.3 | 96.6 | 44.8 | 100.0 | 93.1 |
| Life sciences | bacteria (1) | 24 | 95.5 | 95.8 | 41.7 | 95.8 | 91.7 |
| | Drug (1) | 77 | 98.6 | 98.7 | 39.0 | 94.8 | 94.8 |
| | Enzyme (1) | 56 | 100.0 | 100.0 | 8.9 | 75.0 | 80.4 |
| | microorganism (1) | 15 | 60.0 | 60.0 | 6.7 | 6.7 | 6.7 |
| Mechanics, engineering | Bicycle (3) | 180 | 87.5 | 87.8 | 19.4 | 51.1 | 47.2 |
| | car (1) | 83 | 100.0 | 100.0 | 48.2 | 100.0 | 95.2 |
| | Engine (1) | 31 | 100.0 | 100.0 | 54.8 | 100.0 | 90.3 |
| | Hand tool (1) | 63 | 38.3 | 39.7 | 4.8 | 39.7 | 33.3 |
| | Machine (2) | 1584 | 86.4 | 86.4 | 35.2 | 67.0 | 61.6 |
| | Motor (2) | 76 | 98.6 | 98.7 | 52.6 | 96.1 | 89.5 |
| | saw (1) | 28 | 96.0 | 96.4 | 57.1 | 100.0 | 100.0 |
| | Vehicle (2) | 348 | 89.2 | 84.2 | 29.6 | 66.1 | 63.8 |
| | Wheel (2) | 596 | 83.7 | 84.6 | 51.3 | 76.7 | 73.3 |
| Physics | Instrument (1) | 105 | 85.6 | 85.7 | 50.5 | 61.9 | 50.5 |
| | Light (1) | 102 | 100.0 | 100.0 | 29.4 | 88.2 | 84.3 |
| | Liquid (1) | 29 | 93.1 | 93.1 | 27.6 | 58.6 | 69.0 |
| | Material (1) | 301 | 100.0 | 100.0 | 15.0 | 50.2 | 33.2 |
| | metal (1) | 273 | 95.9 | 96.0 | 37.4 | 71.8 | 71.1 |
| | sensor (1) | 8 | 100.0 | 100.0 | 87.5 | 100.0 | 100.0 |
| | Wave (1) | 71 | 97.2 | 97.2 | 32.4 | 91.5 | 87.3 |
| | Total | 5985 | | | | | |
| | Average percentages | | 90.5 | 90.3 | 43.8 | 74.3 | 70.8 |

Table 4. Percentages of verified hierarchical relations according to scope values calculated using several probability models.

However, these shortcomings can be mitigated by providing context, i.e., by placing the same contextual words before the terms to be compared. For example, the query *computing device* can be replaced by *a computerized system including a computing device* (emphasis put on the added context). Similarly, a *hand tool* can be replaced by *a workshop including a hand tool*. Doing so yields remarkable improvements to the GPT2 results but does obviously not impact the results from the simpler models, unless by chance.

| | | | Percentages of relations verified | | | | |
|---|---|---|---|---|---|---|---|
| Category | Root | Number of tested relations | GPT2 | Word frequency | Word count | Character frequency | Character count |
| ICT | Computer (without context) | 25 | 95.83 | 96.00 | 36.00 | 56.00 | 60.00 |
| | Computer (with context) | 25 | 100.00 | 96.00 | 36.00 | 56.00 | 60.00 |
| | Computing device (without context) | 21 | 25.00 | 28.57 | 0.00 | 9.52 | 9.52 |
| | Computing device (with context) | 21 | 90.48 | 28.57 | 0.00 | 9.52 | 9.52 |
| | Information processing system (without context) | 21 | 15.00 | 4.76 | 0.00 | 0.00 | 0.00 |
| | Information processing system (with context) | 21 | 61.90 | 4.76 | 0.00 | 0.00 | 0.00 |
| Mechanics, engineering | Hand tool (without context) | 63 | 38.3 | 39.7 | 4.8 | 39.7 | 33.3 |
| | Hand tool (with context) | 63 | 81.0 | 39.7 | 4.8 | 39.7 | 33.3 |
| | Wheel (without context) | 48 | 95.5 | 95.8 | 50.0 | 93.8 | 91.7 |
| | Wheel (with context) | 48 | 100.0 | 97.9 | 50.0 | 93.8 | 91.7 |

Table 5. Comparison of percentages of verified hierarchical relations calculated with or without context.

The probabilities obtained for patent claims are similarly impacted by the context set by the claim language. Thus, one may expect LLMs to partly compensate for functional definitions, which are abundantly used in patent claims.

Further factors that impact the results are the depth and breadth of the trees and the various meanings of the compared terms. The larger the tree, the higher the chance for improper comparisons. E.g., a *conductor* has both a technical meaning (referring to an electrical or thermal conductor) and a non-technical meaning (a person who directs the performance of a choir or an orchestra). Now, the dictionary used to identify the hierarchies does not resolve the contexts, which affects the results. This is illustrated in Table 6, which shows results obtained for large graphs constructed from general root terms. That said, the results obtained are still fairly good (> 80 %) for the GPT2 model and the model based on word frequencies, and, to a lesser extent, the character-based models.

All in all, a relation can be established between the breadth of terms and their likeliness. The more probable a term, the broader it is, on average. And, conversely, the less probable a term, the narrower it is, on average. Such a conclusion holds very well for the GPT2 and frequency-based models, as long as single words are compared. Circumlocutions muddles the issue. However, this problem is partly alleviated by the fact that LLMs take into account the context set by other words in a claim. Also, compensation phenomena occur, which benefit the results.

| Category | Root | Number of tested relations | Percentages of relations verified | | | | |
|---|---|---|---|---|---|---|---|
| | | | GPT2 | Word frequency | Word count | Character frequency | Character count |
| General | Conductor | 116723 | 90.9 | 91.2 | 34.7 | 78.7 | 72.9 |
| | Machine | 50695 | 78.3 | 78.2 | 33.0 | 63.5 | 59.9 |
| | System | 4717 | 91.1 | 91.1 | 35.8 | 65.3 | 60.5 |
| | Device | 6281 | 83.0 | 82.9 | 29.9 | 64.1 | 58.9 |
| | Total/average | 185439 | 83.7 | 83.9 | 32.8 | 70.7 | 65.8 |

Table 6. Comparison of percentages of verified hierarchical relations calculated for large graphs.

# 5. Assessment and comparison of the various probability models

The results obtained from the various probability models tested somehow match expectations, insofar as the predicted scope values generally decrease as the number or complexity of the claimed features increases. Furthermore, tight correlations are observed between the results obtained from the various language models. However, interesting differences can also be noted. Overall, the model based on word counts is the least satisfactory. Using instead the character count, which is just as easy to implement, offers substantially better results. In particular, character counts yield lower standard deviations for claims of identical scopes, see Table 3, and better align with known hierarchical relations of breadths of terms, see Tables 5 – 7. This is a fortunate consequence of the fact that longer words have more conceptual complexity and are rarer, on average.

Frequency-based models add some complexity in terms of implementation, something that pays off at times. Most notably, frequency-based models can always discriminate between claims of identical word and character counts, see Section 4.4. The model based on word frequencies slightly improves on the model based on character frequencies, notably in terms of verified hierarchical relations of breadths, as discussed in Section 4.5.

The GPT2 model systematically outperforms simple models based on counts or frequencies of words or characters, as a result of the fact that this model takes into account the context established by the terminologies used in each claim. Interestingly, however, results obtained from larger LLMs do not systematically improve on the GPT2 results, see Table 4 and FIG. 5, even though they improve on the simpler models. This, however, may result from different cutoffs used to compute the probabilities, as noted in section 3.4.1.

# 6. Concluding remarks

This work illustrates how language models can be leveraged to measure the scope of patent claims. The scope $S(C)$ is formulated as the reciprocal of self-information $I(C)$, which is the negative of the logarithm of the probability $p(C)$ associated with a given claim $C$. The probability $p(C)$ is an estimate of the likeliness of claim $C$. This probability is built as a product of probability terms associated with tokens of the claim. Alternatively, it can be formulated as a weighted sum of products of probability terms to better address disclaimers and alternatives, if any.

The probability terms are estimated thanks to language models. Seven language models were tested in this work, ranging from simplistic models (each word or character is drawn from a uniform distribution) to intermediate models (based on average word or character frequencies), to large language models (GPT-2, babbage-002, and davinci-002).

The proposed approach is grounded in the assumption that rarer concepts are more surprising (they are more informative in the sense of information theory) and thus result in a narrower scope. This assumption is verified in the vast majority of cases when comparing single words, less so when comparing terms with different word counts. Fortunately, LLMs partly solve this problem thanks to the context set by the claim language.

When formulating $p(C)$ as a single product of token probabilities, the scope can be shown to be proportional to the reciprocal of the number $N$ of tokens involved in the claims, times a modulating factor. In general, this factor varies rather slowly from one claim to the other; it is even constant where tokens are drawn from uniform distributions. In that case, the scope $S(C)$ becomes proportional to $1/N$, where $N$ can for instance be chosen to correspond to the number of words or characters. Thus, the proposed definition of the claim scope generalizes previous approaches based on word or character counts.

Application was made to nine series of claims directed to distinct inventions, where the claims have a gradually decreasing scope. This property is generally well reflected by the calculated scope values, essentially because the claims happen to have increasing word counts in each series. However, this property is no longer verified systematically for claims having a decreasing scope despite a constant word count, at least when using other models than GPT2. Another limitation is that all language models fail to correctly predict the same scope value for variations on claims of the exact same scope. All in all, the LLMs outperform models based on word or character frequencies, which are themselves ahead of models based on word and character counts.

Interestingly, the character counts yield better results than word counts —a fortunate consequence of the fact that longer words usually are rarer and have more conceptual complexity. Thus, it may be more judicious to rely on character counts than on word counts, at least for claims drafted in the English language. Using character counts may further ensure a better consistency from one language to the other. For instance, comparing the three official versions (English, French, and German) of the examination guidelines of the European Patent Office [3] shows that the standard deviation of the word counts between the three versions is approximately equal to 10% of the mean word count, while this figure falls to approximately 1% when considering characters instead of words.

The good relative performance of the LLMs deserves additional comments. Here, the goal is not to generate text, as generative tools like ChatGPT do, but to estimate, *a posteriori*, the probability with which an actual patent claim would have been completed. Thus, the usual flaws of generative tools (e.g., hallucinations and other alignment issues) do not apply here, at least not in the same way as for text generation. Rather, the limitations of the LLMs used in this work may put into question the probabilities obtained for the claims.

Some issues were noted regarding disclaimers and alternatives, which require further work. Since the proposed scope function depends primarily on $1/N$, all probability models considered herein are invariably impacted by the number $N$ of tokens, which sometimes leads to inconsistencies. In particular, functional definitions and other types of circumlocutions may inadvertently cause to narrow down the calculated scope, whereas they are actually intended to broaden the definitions of the claimed concepts. Such issues are partly solved by the LLMs, thanks to the context set by the claim.

This suggests testing higher-performance LLMs, provided that such models are open (unlike GPT-3.5 and GPT-4) and allow token log-probabilities to be extracted from prompts (e.g., in completion mode) with sufficient accuracy. While larger general models are unlikely to provide decisive improvements, significant progress can be expected from models trained specifically on patent claim language. Efforts are going in this direction [34, 35]. In particular, such models may help alleviate noise generated by synonymous terminologies (e.g., *comprising* vs. *including*), functional definitions, and other circumlocutions.

Automatic scope value calculations would further benefit from systematic normalization procedures. This is desirable as the diversity of patent drafting practices and cultures results in a variety of patent claim formulation styles, something that will inevitably impact the calculated scope values, even when using LLMs. Still, some of these issues can be addressed through text normalization. In particular, NLP can be used to rework the claims according to a given standard [36, 37], before automatically computing the scope of the normalized claims. Another possibility would be to prompt an LLM to reformulate the claims according to a given standard.

Obviously, the results obtained with the present scope functions should be compared with human assessments. However, doing this first requires establishing a suitable protocol, if only to mitigate the inevitable subjectivity of human assessments [38].

Finally, among other questions raised by the growing use of LLMs [24], also for IP matters [39, 40, 41, 42], it is worth recalling limitations inherent to machine learning models [43, 44, 45, 46]. Such models remain statistical models, which may occasionally fail. A trained model does, by definition, infer results that are statistically "normal" in view of data on which it was trained. Thus, the inferences made are satisfactory, on average, but only as long as the tested data remain commensurate with the training data. Now, the language of the data tested (patent claims) in this work differ from that of the training data (e.g., text scraped from web pages), hence the need for models specifically trained on patent claim language.

Clearly, no computational model can capture the depth of human analyses to assess the scope of patent claims. However, when it comes to automatically quantifying the scope of patent claims, a mathematical framework is needed, together with practical calculation schemes, as tentatively proposed in this work.

# Appendix A: Series or claims used for calculations of scope values

Sections A.1 to A.9 list nine series of claims of gradually decreasing scope. The claim references are referred to in Appendix B, listing corresponding scope values, calculated from various probability models.

## A.1. Series of claims to pencil eraser

| Ref. | Claims |
|---|---|
| A1.1 | Writing instrument. |
| A1.2 | Writing instrument comprising a pencil. |
| A1.3 | Writing instrument comprising a pencil and an eraser. |
| A1.4 | Writing instrument comprising a pencil and an eraser, wherein the eraser is attached to the pencil. |
| A1.5 | Writing instrument comprising a pencil and an eraser, wherein the eraser is attached to the pencil through a ferrule. |
| A1.6 | Writing instrument comprising a pencil and an eraser, wherein the eraser is attached to the pencil through a ferrule made as a hollow cylinder. |
| A1.7 | Writing instrument comprising a pencil and an eraser, wherein the eraser is attached to the pencil through a ferrule made as a hollow cylinder comprising aluminum. |
| A1.8 | Writing instrument comprising a pencil and an eraser, wherein the eraser is attached to the pencil through a ferrule made as a hollow cylinder comprising aluminum, and wherein the ferrule has a plurality of serrations. |
| A1.9 | Writing instrument comprising a pencil and an eraser, wherein the eraser is attached to the pencil through a ferrule made as a hollow cylinder comprising aluminum, and wherein the ferrule has a plurality of serrations on an outer surface thereof. |
| A1.10 | Writing instrument comprising a pencil and an eraser, wherein the eraser is attached to the pencil through a ferrule made as a hollow cylinder comprising aluminum, and wherein the ferrule has a plurality of serrations on an outer surface thereof, the serrations covering a major portion of the outer surface of the hollow cylinder. |
| A1.11 | Writing instrument comprising a pencil and an eraser, wherein the eraser is attached to the pencil through a ferrule made as a hollow cylinder comprising aluminum, and wherein the ferrule has a plurality of serrations on an outer surface thereof, the serrations covering a major portion of the outer surface of the hollow cylinder and extending longitudinally over said outer surface. |
| A1.12 | Writing instrument comprising a pencil and an eraser, wherein the eraser is attached to the pencil through a ferrule made as a hollow cylinder comprising aluminum, wherein the ferrule has a plurality of serrations on an outer surface thereof, the serrations covering a major portion of the outer surface of the hollow cylinder and extending longitudinally over said outer surface, and wherein the pencil includes softwood. |
| A1.13 | Writing instrument comprising a pencil and an eraser, wherein the eraser is attached to the pencil through a ferrule made as a hollow cylinder comprising aluminum, wherein the ferrule has a plurality of serrations on an outer surface thereof, the serrations covering a major portion of the outer surface of the hollow cylinder and extending longitudinally over said outer surface, and wherein the pencil includes softwood encasing a solid pigment. |
| A1.14 | Writing instrument comprising a pencil and an eraser, wherein the eraser is attached to the pencil through a ferrule made as a hollow cylinder comprising aluminum, wherein the ferrule has a plurality of serrations on an outer surface thereof, the serrations covering a major portion of the outer surface of the hollow cylinder and extending longitudinally over said outer surface, and wherein the pencil includes softwood encasing a solid pigment, the solid pigment including a powder compressed into a solid core. |
| A1.15 | Writing instrument comprising a pencil and an eraser, wherein the eraser is attached to the pencil through a ferrule made as a hollow cylinder comprising aluminum, wherein the ferrule has a plurality of serrations on an outer surface thereof, the serrations covering a major portion of the outer surface of the hollow cylinder and extending longitudinally over said outer surface, wherein the pencil includes softwood encasing a solid pigment, the solid pigment including a powder compressed into a solid core, and wherein the solid core includes a mixture of graphite, clay, and a binder. |
| A1.16 | Writing instrument comprising a pencil and an eraser, wherein the eraser is attached to the pencil through a ferrule made as a hollow cylinder comprising aluminum, wherein the ferrule has a plurality of serrations on an outer surface thereof, the serrations covering a major portion of the outer surface of the hollow cylinder and extending longitudinally over said outer surface, wherein the pencil includes softwood encasing a solid pigment, the solid pigment including a powder compressed into a solid core, and wherein the solid core includes a mixture of graphite, clay, and a binder, the relative amounts of which are of between 64 and 66%, 29 and 31%, and 3 and 7% by weight, respectively. |
| A1.17 | Writing instrument comprising a pencil and an eraser, wherein the eraser is attached to the pencil through a ferrule made as a hollow cylinder comprising aluminum, wherein the ferrule has a plurality of serrations on an outer surface thereof, the serrations covering a major portion of the outer surface of the hollow cylinder and extending longitudinally over said outer surface, wherein the pencil includes softwood encasing a solid pigment, the solid pigment including a powder compressed into a solid core, wherein the solid core includes a mixture of graphite, clay, and a binder, the relative amounts of which are of between 64 and 66%, 29 and 31%, and 3 and 7% by weight, respectively, and wherein the solid core further comprises alkali metal phosphate. |
| A1.18 | Writing instrument comprising a pencil and an eraser, wherein the eraser is attached to the pencil through a ferrule made as a hollow cylinder comprising aluminum, wherein the ferrule has a plurality of serrations on an outer surface thereof, the serrations covering a major portion of the outer surface of the hollow cylinder and extending longitudinally over said outer surface, wherein the pencil includes softwood encasing a solid pigment, the solid pigment including a powder compressed into a solid core, wherein the solid core includes a mixture of graphite, clay, and a binder, the relative amounts of which are of between 64 and 66%, 29 and 31%, and 3 and 7% by weight, respectively, and wherein the solid core further comprises alkali metal phosphate and alkali metal carbonate. |

| Ref. | Claims |
|---|---|
| A1.19 | Writing instrument comprising a pencil and an eraser, wherein the eraser is attached to the pencil through a ferrule made as a hollow cylinder comprising aluminum, wherein the ferrule has a plurality of serrations on an outer surface thereof, the serrations covering a major portion of the outer surface of the hollow cylinder and extending longitudinally over said outer surface, wherein the pencil includes softwood encasing a solid pigment, the solid pigment including a powder compressed into a solid core, wherein the solid core includes a mixture of graphite, clay, and a binder, the relative amounts of which are of between 64 and 66%, 29 and 31%, and 3 and 7% by weight, respectively, and wherein the solid core further comprises alkali metal phosphate, alkali metal borate, and alkali metal carbonate. |
| A1.20 | Writing instrument comprising a pencil and an eraser, wherein the eraser is attached to the pencil through a ferrule made as a hollow cylinder comprising aluminum, wherein the ferrule has a plurality of serrations on an outer surface thereof, the serrations covering a major portion of the outer surface of the hollow cylinder and extending longitudinally over said outer surface, wherein the pencil includes softwood encasing a solid pigment, the solid pigment including a powder compressed into a solid core, wherein the solid core includes a mixture of graphite, clay, and a binder, the relative amounts of which are of between 64 and 66%, 29 and 31%, and 3 and 7% by weight, respectively, and wherein the solid core further comprises alkali metal phosphate, alkali metal borate, and alkali metal carbonate, in a total amount that is between 0.000001 and 0.100000 % by weight. |
| A1.21 | Writing instrument comprising a pencil and an eraser, wherein the eraser is attached to the pencil through a ferrule made as a hollow cylinder comprising aluminum, wherein the ferrule has a plurality of serrations on an outer surface thereof, the serrations covering a major portion of the outer surface of the hollow cylinder and extending longitudinally over said outer surface, wherein the pencil includes softwood encasing a solid pigment, the solid pigment including a powder compressed into a solid core, wherein the solid core includes a mixture of graphite, clay, and a binder, the relative amounts of which are of between 64 and 66%, 29 and 31%, and 3 and 7% by weight, respectively, wherein the solid core further comprises alkali metal phosphate, alkali metal borate, and alkali metal carbonate, in a total amount that is between 0.000001 and 0.100000 % by weight, and wherein the softwood consists of cedarwood. |
| A1.22 | Writing instrument comprising a pencil and an eraser, wherein the eraser is attached to the pencil through a ferrule made as a hollow cylinder comprising aluminum, wherein the ferrule has a plurality of serrations on an outer surface thereof, the serrations covering a major portion of the outer surface of the hollow cylinder and extending longitudinally over said outer surface, wherein the pencil includes softwood encasing a solid pigment, the solid pigment including a powder compressed into a solid core, wherein the solid core includes a mixture of graphite, clay, and a binder, the relative amounts of which are of between 64 and 66%, 29 and 31%, and 3 and 7% by weight, respectively, wherein the solid core further comprises alkali metal phosphate, alkali metal borate, and alkali metal carbonate, in a total amount that is between 0.000001 and 0.100000 % by weight, and wherein the softwood consists of cedarwood having a density of between 300 and 400 kg/m3. |

## A.2. Series of claims to sensor package

| Ref. | Claim |
|---|---|
| A2.1 | Sensor package, comprising a sensor chip including a sensitive element. |
| A2.2 | Sensor package, comprising a sensor chip including a sensitive element exposed to an environment of the sensor package. |
| A2.3 | Sensor package, comprising a sensor chip including a sensitive element exposed to an environment of the sensor package, the sensitive element being sensitive to an environmental measurand. |
| A2.4 | Sensor package, comprising: a sensor chip including a sensitive element exposed to an environment of the sensor package, the sensitive element being sensitive to an environmental measurand; and contact pads for electrically contacting the sensor package. |
| A2.5 | Sensor package, comprising: a sensor chip including a sensitive element exposed to an environment of the sensor package, the sensitive element being sensitive to an environmental measurand; and contact pads for electrically contacting the sensor package, wherein each of the contact pads is at least partly exposed to the environment. |
| A2.6 | Sensor package, comprising: a sensor chip including a sensitive element exposed to an environment of the sensor package, the sensitive element being sensitive to an environmental measurand; and contact pads for electrically contacting the sensor package, wherein each of the contact pads is at least partly exposed to the environment, so as to be accessible from the outside of the sensor package. |
| A2.7 | Sensor package, comprising: a sensor chip including a sensitive element exposed to an environment of the sensor package, the sensitive element being sensitive to an environmental measurand; contact pads for electrically contacting the sensor package, wherein each of the contact pads is at least partly exposed to the environment, so as to be accessible from the outside of the sensor package; and electrical connections. |
| A2.8 | Sensor package, comprising: a sensor chip including a sensitive element exposed to an environment of the sensor package, the sensitive element being sensitive to an environmental measurand; contact pads for electrically contacting the sensor package, wherein each of the contact pads is at least partly exposed to the environment, so as to be accessible from the outside of the sensor package; and electrical connections between the sensor chip and the contact pads. |
| A2.9 | Sensor package, comprising: a sensor chip including a sensitive element exposed to an environment of the sensor package, the sensitive element being sensitive to an environmental measurand; contact pads for electrically contacting the sensor package, wherein each of the contact pads is at least partly exposed to the environment, so as to be accessible from the outside of the sensor package; and electrical connections between the sensor chip and the contact pads, the electrical connections being bond wires. |
| A2.10 | Sensor package, comprising: a sensor chip including a sensitive element exposed to an environment of the sensor package, the sensitive element being sensitive to an environmental measurand; contact pads for electrically contacting the sensor package, wherein each of the contact pads is at least partly exposed to the environment, so as to be accessible from the outside of the sensor package; electrical connections between the sensor chip and the contact pads, the electrical connections being bond wires; and a molding compound. |
| A2.11 | Sensor package, comprising: a sensor chip including a sensitive element exposed to an environment of the sensor package, the sensitive element being sensitive to an environmental measurand; contact pads for electrically contacting the sensor package, wherein each of the contact pads is at least partly exposed to the environment, so as to be accessible from the outside of the sensor package; electrical connections between the sensor chip and the contact pads, the electrical connections being bond wires; and a molding compound at least partially enclosing the sensor chip and the contact pads. |

| Ref. | Claim |
|---|---|
| A2.12 | Sensor package, comprising: a sensor chip including a sensitive element exposed to an environment of the sensor package, the sensitive element being sensitive to an environmental measurand; contact pads for electrically contacting the sensor package, wherein each of the contact pads is at least partly exposed to the environment, so as to be accessible from the outside of the sensor package; electrical connections between the sensor chip and the contact pads, the electrical connections being bond wires; and a molding compound at least partially enclosing the sensor chip and the contact pads, wherein the sensor chip is arranged in the sensor package with respect to the contact pads such that a top surface of the sensor chip does not protrude from a level defined by a top surface of the contact pads. |
| A2.13 | Sensor package, comprising: a sensor chip including a sensitive element exposed to an environment of the sensor package, the sensitive element being sensitive to an environmental measurand; contact pads for electrically contacting the sensor package, wherein each of the contact pads is at least partly exposed to the environment, so as to be accessible from the outside of the sensor package; electrical connections between the sensor chip and the contact pads, the electrical connections being bond wires; and a molding compound at least partially enclosing the sensor chip and the contact pads, wherein the sensor chip is arranged in the sensor package with respect to the contact pads such that a top surface of the sensor chip does not protrude from a level defined by a top surface of the contact pads and a bottom surface of the sensor chip does not protrude from a level defined by a bottom surface of the contact pads. |
| A2.14 | Sensor package, comprising: a sensor chip including a sensitive element exposed to an environment of the sensor package, the sensitive element being sensitive to an environmental measurand; contact pads for electrically contacting the sensor package, wherein each of the contact pads is at least partly exposed to the environment, so as to be accessible from the outside of the sensor package; electrical connections between the sensor chip and the contact pads, the electrical connections being bond wires; and a molding compound at least partially enclosing the sensor chip and the contact pads, wherein the sensor chip is arranged in the sensor package with respect to the contact pads such that a top surface of the sensor chip does not protrude from a level defined by a top surface of the contact pads and a bottom surface of the sensor chip does not protrude from a level defined by a bottom surface of the contact pads, and such that an extension of the sensor package is limited by levels defined by the top and the bottom surfaces of the contact pads. |
| A2.15 | Sensor package, comprising: a sensor chip including a sensitive element exposed to an environment of the sensor package, the sensitive element being sensitive to an environmental measurand; contact pads for electrically contacting the sensor package, wherein each of the contact pads is at least partly exposed to the environment, so as to be accessible from the outside of the sensor package; electrical connections between the sensor chip and the contact pads, the electrical connections being bond wires; and a molding compound at least partially enclosing the sensor chip and the contact pads, wherein the sensor chip is arranged in the sensor package with respect to the contact pads such that a top surface of the sensor chip does not protrude from a level defined by a top surface of the contact pads and a bottom surface of the sensor chip does not protrude from a level defined by a bottom surface of the contact pads, and such that an extension of the sensor package, as measured along a direction z, is limited by levels defined by the top and the bottom surfaces of the contact pads, and wherein said direction z is orthogonal to each of the top and bottom surfaces of the sensor chip and the contact pads. |
| A2.16 | Sensor package, comprising: a sensor chip including a sensitive element exposed to an environment of the sensor package, the sensitive element being sensitive to an environmental measurand; contact pads for electrically contacting the sensor package, wherein each of the contact pads is at least partly exposed to the environment, so as to be accessible from the outside of the sensor package; electrical connections between the sensor chip and the contact pads, the electrical connections being bond wires; and a molding compound at least partially enclosing the sensor chip and the contact pads, wherein the sensor chip is arranged in the sensor package with respect to the contact pads such that a top surface of the sensor chip does not protrude from a level defined by a top surface of the contact pads and a bottom surface of the sensor chip does not protrude from a level defined by a bottom surface of the contact pads, and such that an extension of the sensor package, as measured along a direction z, is limited by levels defined by the top and the bottom surfaces of the contact pads, wherein said direction z is orthogonal to each of the top and bottom surfaces of the sensor chip and the contact pads, and wherein each of the contact pads is exposed from both a bottom surface and a top surface of the sensor package, the top surface of the sensor package being opposite to the bottom surface of the sensor package. |
| A2.17 | Sensor package, comprising: a sensor chip including a sensitive element exposed to an environment of the sensor package, the sensitive element being sensitive to an environmental measurand; contact pads for electrically contacting the sensor package, wherein each of the contact pads is at least partly exposed to the environment, so as to be accessible from the outside of the sensor package; electrical connections between the sensor chip and the contact pads, the electrical connections being bond wires; and a molding compound at least partially enclosing the sensor chip and the contact pads, wherein the sensor chip is arranged in the sensor package with respect to the contact pads such that a top surface of the sensor chip does not protrude from a level defined by a top surface of the contact pads and a bottom surface of the sensor chip does not protrude from a level defined by a bottom surface of the contact pads, and such that an extension of the sensor package, as measured along a direction z, is limited by levels defined by the top and the bottom surfaces of the contact pads, wherein said direction z is orthogonal to each of the top and bottom surfaces of the sensor chip and the contact pads, wherein each of the contact pads is exposed from both a bottom surface and a top surface of the sensor package, the top surface of the sensor package being opposite to the bottom surface of the sensor package, and wherein each contact pad has a landing for the assigned electrical connection at a level below the top surface of the contact pad and arranged on a side facing a side face of the sensor chip. |
| A2.18 | Sensor package, comprising: a sensor chip including a sensitive element exposed to an environment of the sensor package, the sensitive element being sensitive to an environmental measurand; contact pads for electrically contacting the sensor package, wherein each of the contact pads is at least partly exposed to the environment, so as to be accessible from the outside of the sensor package; electrical connections between the sensor chip and the contact pads, the electrical connections being bond wires; and a molding compound at least partially enclosing the sensor chip and the contact pads, wherein the sensor chip is arranged in the sensor package with respect to the contact pads such that a top surface of the sensor chip does not protrude from a level defined by a top surface of the contact pads and a bottom surface of the sensor chip does not protrude from a level defined by a bottom surface of the contact pads, and such that an extension of the sensor package, as measured along a direction z, is limited by levels defined by the top and the bottom surfaces of the contact pads, wherein said direction z is orthogonal to each of the top and bottom surfaces of the sensor chip and the contact pads, wherein each of the contact pads is exposed from both a bottom surface and a top surface of the sensor package, the top surface of the sensor package being opposite to the bottom surface of the sensor package, and wherein each contact pad has a landing for the assigned electrical connection at a level below the top surface of the contact pad and arranged on a side facing a side face of the sensor chip, and wherein the sensor chip has terminals at its top surface for the electrical connections. |

| Ref. | Claim |
|---|---|
| A2.19 | Sensor package, comprising: a sensor chip including a sensitive element exposed to an environment of the sensor package, the sensitive element being sensitive to an environmental measurand; contact pads for electrically contacting the sensor package, wherein each of the contact pads is at least partly exposed to the environment, so as to be accessible from the outside of the sensor package; electrical connections between the sensor chip and the contact pads, the electrical connections being bond wires; and a molding compound at least partially enclosing the sensor chip and the contact pads, wherein the sensor chip is arranged in the sensor package with respect to the contact pads such that a top surface of the sensor chip does not protrude from a level defined by a top surface of the contact pads and a bottom surface of the sensor chip does not protrude from a level defined by a bottom surface of the contact pads, and such that an extension of the sensor package, as measured along a direction z, is limited by levels defined by the top and the bottom surfaces of the contact pads, wherein said direction z is orthogonal to each of the top and bottom surfaces of the sensor chip and the contact pads, wherein each of the contact pads is exposed from both a bottom surface and a top surface of the sensor package, the top surface of the sensor package being opposite to the bottom surface of the sensor package, and wherein each contact pad has a landing for the assigned electrical connection at a level below the top surface of the contact pad and arranged on a side facing a side face of the sensor chip, wherein the sensor chip has terminals at its top surface for the electrical connections, and wherein the molding compound at least fills a space between the contact pads and the sensor chip and embeds the electrical connections. |

## A.3. Series of claims to optical device

| Ref. | Claim |
|---|---|
| A3.1 | Optical device. |
| A3.2 | Optical device comprising a layer structure. |
| A3.3 | Optical device comprising a layer structure, which includes a thermally conducting optical reflector. |
| A3.4 | Optical device comprising a layer structure, the layer structure including a thermally conducting optical reflector; and a thermally conducting spacer. |
| A3.5 | Optical device comprising a layer structure, the layer structure including a thermally conducting optical reflector; and a thermally conducting spacer, which is transmissive to light. |
| A3.6 | Optical device comprising a layer structure, the layer structure including a thermally conducting optical reflector; and a thermally conducting spacer, which is transmissive to light and arranged above the reflector. |
| A3.7 | Optical device comprising a layer structure, the layer structure including: a thermally conducting optical reflector; a thermally conducting spacer, which is transmissive to light and arranged above the reflector; and a phase-change material arranged above the spacer. |
| A3.8 | Optical device comprising a layer structure, the layer structure including: a thermally conducting optical reflector; a thermally conducting spacer, which is transmissive to light and arranged above the reflector; and a phase-change material arranged above the spacer, wherein the phase-change material has at least two reversibly switchable states. |
| A3.9 | Optical device comprising a layer structure, the layer structure including: a thermally conducting optical reflector; a thermally conducting spacer, which is transmissive to light and arranged above the reflector; and a phase-change material arranged above the spacer, wherein the phase-change material has at least two reversibly switchable states, in which the phase-change material exhibits two different values of refractive index. |
| A3.10 | Optical device comprising a layer structure, the layer structure including: a thermally conducting optical reflector; a thermally conducting spacer, which is transmissive to light and arranged above the reflector; and a phase-change material arranged above the spacer, wherein the phase-change material has at least two reversibly switchable states, in which the phase-change material exhibits two different values of refractive index, wherein the reflector, the spacer, and the phase-change material, are successively stacked along a stacking direction of the layer structure. |
| A3.11 | Optical device comprising a layer structure, the layer structure including: a thermally conducting optical reflector; a thermally conducting spacer, which is transmissive to light and arranged above the reflector; and a phase-change material arranged above the spacer, wherein the phase-change material has at least two reversibly switchable states, in which the phase-change material exhibits two different values of refractive index, wherein the reflector, the spacer, and the phase-change material, are successively stacked along a stacking direction of the layer structure, wherein the optical device further comprises a heating element opposite to the phase-change material with respect to the reflector. |
| A3.12 | Optical device comprising a layer structure, the layer structure including: a thermally conducting optical reflector; a thermally conducting spacer, which is transmissive to light and arranged above the reflector; and a phase-change material arranged above the spacer, wherein the phase-change material has at least two reversibly switchable states, in which the phase-change material exhibits two different values of refractive index, wherein the reflector, the spacer, and the phase-change material, are successively stacked along a stacking direction of the layer structure, wherein the optical device further comprises a heating element opposite to the phase-change material with respect to the reflector, wherein the layer structure is configured to electrically insulate the phase-change material from the heating element. |
| A3.13 | Optical device comprising a layer structure, the layer structure including: a thermally conducting optical reflector; a thermally conducting spacer, which is transmissive to light and arranged above the reflector; and a phase-change material arranged above the spacer, wherein the phase-change material has at least two reversibly switchable states, in which the phase-change material exhibits two different values of refractive index, wherein the reflector, the spacer, and the phase-change material, are successively stacked along a stacking direction of the layer structure, wherein the optical device further comprises a heating element opposite to the phase-change material with respect to the reflector, wherein the layer structure is configured to electrically insulate the phase-change material from the heating element while the heating element is in thermal communication with the phase-change material through the reflector and the spacer. |
| A3.14 | Optical device comprising a layer structure, the layer structure including: a thermally conducting optical reflector; a thermally conducting spacer, which is transmissive to light and arranged above the reflector; and a phase-change material arranged above the spacer, wherein the phase-change material has at least two reversibly switchable states, in which the phase-change material exhibits two different values of refractive index, wherein the reflector, the spacer, and the phase-change material, are successively stacked along a stacking direction of the layer structure, wherein the optical device further comprises a heating element opposite to the phase-change material with respect to |

| Ref. | Claim |
|---|---|
| | the reflector, wherein the layer structure is configured to electrically insulate the phase-change material from the heating element while the heating element is in thermal communication with the phase-change material through the reflector and the spacer, and wherein the optical device further comprises a controller configured to energize the heating element to heat the phase-change material. |
| A3.15 | Optical device comprising a layer structure, the layer structure including: a thermally conducting optical reflector; a thermally conducting spacer, which is transmissive to light and arranged above the reflector; and a phase-change material arranged above the spacer, wherein the phase-change material has at least two reversibly switchable states, in which the phase-change material exhibits two different values of refractive index, wherein the reflector, the spacer, and the phase-change material, are successively stacked along a stacking direction of the layer structure, wherein the optical device further comprises a heating element opposite to the phase-change material with respect to the reflector, wherein the layer structure is configured to electrically insulate the phase-change material from the heating element while the heating element is in thermal communication with the phase-change material through the reflector and the spacer, and wherein the optical device further comprises a controller configured to energize the heating element to heat the phase-change material and thereby reversibly change a refractive index of said phase-change material. |

## A.4. Series of claims to biodegradable nanoparticles

| Ref. | Claim |
|---|---|
| A4.1 | Biodegradable nanoparticles |
| A4.2 | Biodegradable nanoparticles based on a biodegradable homopolymer |
| A4.3 | Biodegradable nanoparticles based on a biodegradable homopolymer of C2–C6 alkyl 2-cyanoacrylate monomers |
| A4.4 | Biodegradable nanoparticles based on a biodegradable homopolymer of C2–C6 alkyl 2-cyanoacrylate monomers, comprising a non-covalent insulin complex encapsulated therein |
| A4.5 | Biodegradable nanoparticles based on a biodegradable homopolymer of C2–C6 alkyl 2-cyanoacrylate monomers, comprising a non-covalent insulin complex encapsulated therein along with a pharmaceutically acceptable stabilizer |
| A4.6 | Biodegradable nanoparticles based on a biodegradable homopolymer of C2–C6 alkyl 2-cyanoacrylate monomers, comprising a non-covalent insulin complex encapsulated therein along with a pharmaceutically acceptable stabilizer, wherein the nanoparticles have a hydrodynamic diameter of 300 nm or less |
| A4.7 | Biodegradable nanoparticles based on a biodegradable homopolymer of C2–C6 alkyl 2-cyanoacrylate monomers, comprising a non-covalent insulin complex encapsulated therein along with a pharmaceutically acceptable stabilizer, wherein the nanoparticles have a hydrodynamic diameter of 300 nm or less as measured by dynamic light scattering |
| A4.8 | Biodegradable nanoparticles based on a biodegradable homopolymer of C2–C6 alkyl 2-cyanoacrylate monomers, comprising a non-covalent insulin complex encapsulated therein along with a pharmaceutically acceptable stabilizer, wherein the nanoparticles have a hydrodynamic diameter of 300 nm or less as measured by dynamic light scattering and comprise 10 to 30 % by weight of insulin relative to the total weight of the nanoparticles |
| A4.9 | Biodegradable nanoparticles based on a biodegradable homopolymer of C2–C6 alkyl 2-cyanoacrylate monomers, comprising a non-covalent insulin complex encapsulated therein along with a pharmaceutically acceptable stabilizer, wherein at least 90 % of the nanoparticles have a hydrodynamic diameter between 100 nm and 300 nm as measured by dynamic light scattering and comprise 10 to 30 % by weight of insulin relative to the total weight of the nanoparticles |
| A4.10 | Biodegradable nanoparticles based on a biodegradable homopolymer of C2–C6 alkyl 2-cyanoacrylate monomers selected from ethyl 2-cyanoacrylate and n-butyl 2-cyanoacrylate, the nanoparticles comprising a non-covalent insulin complex encapsulated therein along with a pharmaceutically acceptable stabilizer, wherein at least 90 % of the nanoparticles have a hydrodynamic diameter between 100 nm and 300 nm as measured by dynamic light scattering and comprise 10 to 30 % by weight of insulin relative to the total weight of the nanoparticles |
| A4.11 | Biodegradable nanoparticles based on a biodegradable homopolymer of C2–C6 alkyl 2-cyanoacrylate monomers selected from ethyl 2-cyanoacrylate and n-butyl 2-cyanoacrylate, the nanoparticles comprising a non-covalent insulin complex encapsulated therein along with a pharmaceutically acceptable stabilizer selected from dextran, chitosan and pectin, wherein at least 90 % of the nanoparticles have a hydrodynamic diameter between 100 nm and 300 nm as measured by dynamic light scattering and comprise 10 to 30 % by weight of insulin relative to the total weight of the nanoparticles |
| A4.12 | Biodegradable nanoparticles based on a biodegradable homopolymer of C2–C6 alkyl 2-cyanoacrylate monomers selected from ethyl 2-cyanoacrylate and n-butyl 2-cyanoacrylate, the nanoparticles comprising a non-covalent insulin complex encapsulated therein along with a pharmaceutically acceptable stabilizer being dextran, wherein at least 90 % of the nanoparticles have a hydrodynamic diameter between 100 nm and 300 nm as measured by dynamic light scattering and comprise 10 to 30 % by weight of insulin relative to the total weight of the nanoparticles |
| A4.13 | Biodegradable nanoparticles based on a biodegradable homopolymer of C2–C6 alkyl 2-cyanoacrylate monomers selected from ethyl 2-cyanoacrylate and n-butyl 2-cyanoacrylate, the nanoparticles comprising a non-covalent insulin complex encapsulated therein along with a pharmaceutically acceptable stabilizer being chitosan, wherein at least 90 % of the nanoparticles have a hydrodynamic diameter between 100 nm and 300 nm as measured by dynamic light scattering and comprise 10 to 30 % by weight of insulin relative to the total weight of the nanoparticles |
| A4.14 | Biodegradable nanoparticles based on a biodegradable homopolymer of C2–C6 alkyl 2-cyanoacrylate monomers selected from ethyl 2-cyanoacrylate and n-butyl 2-cyanoacrylate, the nanoparticles comprising a non-covalent insulin complex encapsulated therein along with a pharmaceutically acceptable stabilizer being pectin, wherein at least 90 % of the nanoparticles have a hydrodynamic diameter between 100 nm and 300 nm as measured by dynamic light scattering and comprise 10 to 30 % by weight of insulin relative to the total weight of the nanoparticles |

## A.5. Series of claims to datacentric computer-implemented method

| Ref. | Claim |
|---|---|
| A5.1 | A computer-implemented method for managing processing resources of a computerized system |
| A5.2 | A computer-implemented method for managing processing resources of a computerized system having at least a first processor and a second processor |
| A5.3 | A computer-implemented method for managing processing resources of a computerized system having at least a first processor and a second processor, where each of the processors is operatively interconnected to a memory adapted to store a set of data |
| A5.4 | A computer-implemented method for managing processing resources of a computerized system having at least a first processor and a second processor, each of the processors operatively interconnected to a memory adapted to store a set of data, the computer implemented method comprising monitoring data of a set |
| A5.5 | A computer-implemented method for managing processing resources of a computerized system having at least a first processor and a second processor, each of the processors operatively interconnected to a memory adapted to store a set of data, the computer implemented method comprising monitoring data of a set, wherein the data monitored are data processed as input data or output data by the first processor while executing |
| A5.6 | A computer-implemented method for managing processing resources of a computerized system having at least a first processor and a second processor, each of the processors operatively interconnected to a memory adapted to store a set of data, the computer implemented method comprising: monitoring data of a set which are processed as at least one of input data or output data by the first processor while executing; and if, based on the monitoring, the second processor is found to be at a shorter distance than the first processor from the monitored data, instructing to interrupt an execution at the first processor and resume the execution at the second processor |
| A5.7 | A computer-implemented method for managing processing resources of a computerized system having at least a first processor and a second processor, each of the processors operatively interconnected to a memory adapted to store a set of data, the computer implemented method comprising: monitoring data of a set which are processed as at least one of input data or output data by the first processor while executing; and if, based on the monitoring, the second processor is found to be at a shorter distance than the first processor from the monitored data, instructing to interrupt an execution at the first processor and resume the execution at the second processor, starting from a processor state that is at least partly determined by a given processor state of the first processor when interrupting the execution at the first processor |

## A.6. Series of claims to siphon

| Ref. | Claim |
|---|---|
| A6.1 | Siphon comprising a housing and a reservoir. |
| A6.2 | Siphon comprising a housing and a reservoir, the reservoir located within the housing. |
| A6.3 | Siphon comprising a housing and a reservoir, the reservoir located within the housing and arranged to hold liquid up to an overflow level of the reservoir. |
| A6.4 | Siphon comprising a housing and a reservoir, the reservoir located within the housing and arranged to hold liquid up to an overflow level of the reservoir, wherein the housing comprises an inlet opening and an outlet opening arranged such that liquid can flow from the inlet opening to the outlet opening via the reservoir. |
| A6.5 | Siphon comprising a housing and a reservoir, the reservoir located within the housing and arranged to hold liquid up to an overflow level of the reservoir, wherein the housing comprises an inlet opening and an outlet opening arranged such that liquid can flow from the inlet opening to the outlet opening via the reservoir, the siphon further comprising an internal wall within the housing. |
| A6.6 | Siphon comprising a housing and a reservoir, the reservoir located within the housing and arranged to hold liquid up to an overflow level of the reservoir, wherein the housing comprises an inlet opening and an outlet opening arranged such that liquid can flow from the inlet opening to the outlet opening via the reservoir, the siphon further comprising an internal wall within the housing, arranged such that when the reservoir is filled with liquid to the overflow level, an inner side of the internal wall is exposed to gas entering the housing from the inlet opening. |
| A6.7 | Siphon comprising a housing and a reservoir, the reservoir located within the housing and arranged to hold liquid up to an overflow level of the reservoir, wherein the housing comprises an inlet opening and an outlet opening arranged such that liquid can flow from the inlet opening to the outlet opening via the reservoir, the siphon further comprising an internal wall within the housing, arranged such that when the reservoir is filled with liquid to the overflow level, an inner side of the internal wall is exposed to gas entering the housing from the inlet opening and the outer side of the internal wall is exposed to gas entering the housing from the outlet opening, thereby blocking gas from flowing from the outlet opening to the inlet opening. |
| A6.8 | Siphon comprising a housing and a reservoir, the reservoir located within the housing and arranged to hold liquid up to an overflow level of the reservoir, wherein the housing comprises an inlet opening and an outlet opening arranged such that liquid can flow from the inlet opening to the outlet opening via the reservoir, the siphon further comprising an internal wall within the housing, arranged such that when the reservoir is filled with liquid to the overflow level, an inner side of the internal wall is exposed to gas entering the housing from the inlet opening and the outer side of the internal wall is exposed to gas entering the housing from the outlet opening, thereby blocking gas from flowing from the outlet opening to the inlet opening, and wherein the internal wall is a tubular wall with one end of the tubular wall surrounding the inlet opening. |
| A6.9 | Siphon comprising a housing and a reservoir, the reservoir located within the housing and arranged to hold liquid up to an overflow level of the reservoir, wherein the housing comprises an inlet opening and an outlet opening arranged such that liquid can flow from the inlet opening to the outlet opening via the reservoir, the siphon further comprising an internal wall within the housing, arranged such that when the reservoir is filled with liquid to the overflow level, an inner side of the internal wall is exposed to gas entering the housing from the inlet opening and the outer side of the internal wall is exposed to gas entering the housing from the outlet opening, thereby blocking |

| Ref. | Claim |
|---|---|
| | gas from flowing from the outlet opening to the inlet opening, wherein the internal wall is a tubular wall with one end of the tubular wall surrounding the inlet opening, and wherein the tubular wall is integrally formed with an inlet tube extending from the housing and for connecting to a sink. |
| A6.10 | Siphon comprising a housing and a reservoir, the reservoir located within the housing and arranged to hold liquid up to an overflow level of the reservoir, wherein the housing comprises an inlet opening and an outlet opening arranged such that liquid can flow from the inlet opening to the outlet opening via the reservoir, the siphon further comprising an internal wall within the housing, arranged such that when the reservoir is filled with liquid to the overflow level, an inner side of the internal wall is exposed to gas entering the housing from the inlet opening and the outer side of the internal wall is exposed to gas entering the housing from the outlet opening, thereby blocking gas from flowing from the outlet opening to the inlet opening, wherein the internal wall is a tubular wall with one end of the tubular wall surrounding the inlet opening, wherein the tubular wall is integrally formed with an inlet tube extending from the housing and for connecting to a sink, and wherein the outlet opening is provided in a side wall of the housing at a height that determines the overflow level. |

## A.7. Series of claims to force sensor

| Ref. | Claim |
|---|---|
| A7.1 | Force sensor. |
| A7.2 | Force sensor for detecting forces applied at multiple locations. |
| A7.3 | Force sensor for detecting forces applied at multiple locations, the force sensor comprising a plurality of input optical fibers. |
| A7.4 | Force sensor for detecting forces applied at multiple locations, the force sensor comprising: a plurality of input optical fibers; and at least one output optical fiber. |
| A7.5 | Force sensor for detecting forces applied at multiple locations, the force sensor comprising: a plurality of input optical fibers; and at least one output optical fiber, each output optical fiber arranged to overlap each input optical fiber at a respective coupling location where light can be coupled from the respective input optical fiber into the respective output optical fiber upon application of a force. |
| A7.6 | Force sensor for detecting forces applied at multiple locations, the force sensor comprising: a plurality of input optical fibers; at least one output optical fiber, each output optical fiber arranged to overlap each input optical fiber at a respective coupling location where light can be coupled from the respective input optical fiber into the respective output optical fiber upon application of a force; and a light injecting device configured to inject an optical signal into each input optical fiber. |
| A7.7 | Force sensor for detecting forces applied at multiple locations, the force sensor comprising: a plurality of input optical fibers; at least one output optical fiber, each output optical fiber arranged to overlap each input optical fiber at a respective coupling location where light can be coupled from the respective input optical fiber into the respective output optical fiber upon application of a force; and a light injecting device configured to inject an optical signal into each input optical fiber, wherein at least two optical signals have at least one different characteristic which allows the at least two optical signals to be distinguished from each other. |
| A7.8 | Force sensor for detecting forces applied at multiple locations, the force sensor comprising: a plurality of input optical fibers; at least one output optical fiber, each output optical fiber arranged to overlap each input optical fiber at a respective coupling location where light can be coupled from the respective input optical fiber into the respective output optical fiber upon application of a force; a light injecting device configured to inject an optical signal into each input optical fiber, wherein at least two optical signals have at least one different characteristic which allows the at least two optical signals to be distinguished from each other; and a light receiving device arranged to detect independently a coupled optical signal from each output optical fiber. |
| A7.9 | Force sensor for detecting forces applied at multiple locations, the force sensor comprising: a plurality of input optical fibers; at least one output optical fiber, each output optical fiber arranged to overlap each input optical fiber at a respective coupling location where light can be coupled from the respective input optical fiber into the respective output optical fiber upon application of a force; a light injecting device configured to inject an optical signal into each input optical fiber, wherein at least two optical signals have at least one different characteristic which allows the at least two optical signals to be distinguished from each other; and a light receiving device arranged to detect independently a coupled optical signal from each output optical fiber, wherein the light receiving device comprises a processing unit configured to determine at least one characteristic of the detected coupled optical signals. |
| A7.10 | Force sensor for detecting forces applied at multiple locations, the force sensor comprising: a plurality of input optical fibers; at least one output optical fiber, each output optical fiber arranged to overlap each input optical fiber at a respective coupling location where light can be coupled from the respective input optical fiber into the respective output optical fiber upon application of a force; a light injecting device configured to inject an optical signal into each input optical fiber, wherein at least two optical signals have at least one different characteristic which allows the at least two optical signals to be distinguished from each other; and a light receiving device arranged to detect independently a coupled optical signal from each output optical fiber, wherein the light receiving device comprises a processing unit configured to determine at least one characteristic of the detected coupled optical signals, corresponding to at least one of the characteristics distinguishing the optical signals injected into the input optical fibers. |
| A7.11 | Force sensor for detecting forces applied at multiple locations, the force sensor comprising: a plurality of input optical fibers; at least one output optical fiber, each output optical fiber arranged to overlap each input optical fiber at a respective coupling location where light can be coupled from the respective input optical fiber into the respective output optical fiber upon application of a force; a light injecting device configured to inject an optical signal into each input optical fiber, wherein at least two optical signals have at least one different characteristic which allows the at least two optical signals to be distinguished from each other; and a light receiving device arranged to detect independently a coupled optical signal from each output optical fiber, wherein the light receiving device comprises a processing unit configured to determine at least one characteristic of the detected coupled optical signals, corresponding to at least one of the characteristics distinguishing the optical signals injected into the input optical fibers, so as to determine a coupling location at which a force is applied. |
| A7.12 | Force sensor for detecting forces applied at multiple locations, the force sensor comprising: a plurality of input optical fibers; at least one output optical fiber, each output optical fiber arranged to overlap each input optical fiber at a respective coupling location where light can |

| Ref. | Claim |
|---|---|
| | be coupled from the respective input optical fiber into the respective output optical fiber upon application of a force; a light injecting device configured to inject an optical signal into each input optical fiber, wherein at least two optical signals have at least one different characteristic which allows the at least two optical signals to be distinguished from each other; and a light receiving device arranged to detect independently a coupled optical signal from each output optical fiber, wherein the light receiving device comprises a processing unit configured to determine at least one characteristic of the detected coupled optical signals, corresponding to at least one of the characteristics distinguishing the optical signals injected into the input optical fibers, so as to determine a coupling location at which a force is applied, wherein at least one of the different characteristics is a different pulse frequency. |
| A7.13 | Force sensor for detecting forces applied at multiple locations, the force sensor comprising: a plurality of input optical fibers; at least one output optical fiber, each output optical fiber arranged to overlap each input optical fiber at a respective coupling location where light can be coupled from the respective input optical fiber into the respective output optical fiber upon application of a force; a light injecting device configured to inject an optical signal into each input optical fiber, wherein at least two optical signals have at least one different characteristic which allows the at least two optical signals to be distinguished from each other; and a light receiving device arranged to detect independently a coupled optical signal from each output optical fiber, wherein the light receiving device comprises a processing unit configured to determine at least one characteristic of the detected coupled optical signals, corresponding to at least one of the characteristics distinguishing the optical signals injected into the input optical fibers, so as to determine a coupling location at which a force is applied, wherein at least one of the different characteristics is a different pulse frequency and the light injecting device comprises a light source and a control unit configured to turn the light source on and off at a particular frequency. |
| A7.14 | Force sensor for detecting forces applied at multiple locations, the force sensor comprising: a plurality of input optical fibers; at least one output optical fiber, each output optical fiber arranged to overlap each input optical fiber at a respective coupling location where light can be coupled from the respective input optical fiber into the respective output optical fiber upon application of a force; a light injecting device configured to inject an optical signal into each input optical fiber, wherein at least two optical signals have at least one different characteristic which allows the at least two optical signals to be distinguished from each other; and a light receiving device arranged to detect independently a coupled optical signal from each output optical fiber, wherein the light receiving device comprises a processing unit configured to determine at least one characteristic of the detected coupled optical signals, corresponding to at least one of the characteristics distinguishing the optical signals injected into the input optical fibers, so as to determine a coupling location at which a force is applied, wherein at least one of the different characteristics is a different pulse frequency, and wherein the light injecting device comprises a light source and a control unit configured to turn the light source on and off at a particular frequency and the light injecting device comprises a shutter with a moveable element which allows or blocks passage of light from a light source into an input fiber, and a control unit configured to open and close the shutter at a particular frequency. |

## A.8. Series of claims to a method of producing a nanoemulsion

| Ref. | Claim |
|---|---|
| A8.1 | Method of producing a nanoemulsion. |
| A8.2 | Method of producing a nanoemulsion, wherein the method comprises mixing an aqueous phase with an oil phase. |
| A8.3 | Method of producing a nanoemulsion, wherein the method comprises: mixing an aqueous phase with an oil phase, the aqueous phase containing a botulinum protein. |
| A8.4 | Method of producing a nanoemulsion, wherein the method comprises: mixing an aqueous phase with an oil phase, the aqueous phase containing a botulinum protein conjugated to polyethylene glycol. |
| A8.5 | Method of producing a nanoemulsion, wherein the method comprises: mixing an aqueous phase with an oil phase, the aqueous phase containing a botulinum protein conjugated to polyethylene glycol having an average molecular weight of 2000 to 15000 Daltons. |
| A8.6 | Method of producing a nanoemulsion, wherein the method comprises: mixing an aqueous phase with an oil phase, the aqueous phase containing a botulinum protein conjugated to polyethylene glycol having an average molecular weight of 2000 to 15000 Daltons; and exposing the mixture to a pressure of at least 1000 bar. |
| A8.7 | Method of producing a nanoemulsion, wherein the method comprises: mixing an aqueous phase with an oil phase, the aqueous phase containing a botulinum protein conjugated to polyethylene glycol having an average molecular weight of 2000 to 15000 Daltons; and exposing the mixture to a pressure of at least 1000 bar for a period of between 30 seconds to 10 minutes. |
| A8.8 | Method of producing a nanoemulsion, wherein the method comprises: mixing an aqueous phase with an oil phase, the aqueous phase containing a botulinum protein conjugated to polyethylene glycol having an average molecular weight of 2000 to 15000 Daltons; and exposing the mixture to a pressure of 1500 to 2000 bar for a period of between 30 seconds to 10 minutes. |
| A8.9 | Method of producing a nanoemulsion, wherein the method comprises: mixing an aqueous phase with an oil phase comprising a surfactant, the aqueous phase containing a botulinum protein conjugated to polyethylene glycol having an average molecular weight of 2000 to 15000 Daltons; and exposing the mixture to a pressure of 1500 to 2000 bar for a period of between 30 seconds to 10 minutes. |
| A8.10 | Method of producing a nanoemulsion, wherein the method comprises: mixing an aqueous phase with an oil phase comprising lecithin, the aqueous phase containing a botulinum protein conjugated to polyethylene glycol having an average molecular weight of 2000 to 15000 Daltons; and exposing the mixture to a pressure of 1500 to 2000 bar for a period of between 30 seconds to 10 minutes. |
| A8.11 | Method of producing a nanoemulsion, wherein the method comprises: mixing an aqueous phase with an oil phase comprising lecithin, the aqueous phase containing a botulinum protein conjugated to polyethylene glycol having an average molecular weight of 2000 to 15000 Daltons; and exposing the mixture to a pressure of 1500 to 2000 bar for a period of between 30 seconds to 10 minutes, wherein the aqueous phase makes up at least 90 volume%. |
| A8.12 | Method of producing a nanoemulsion, wherein the method comprises: mixing an aqueous phase with an oil phase comprising lecithin, the aqueous phase containing a botulinum protein conjugated to polyethylene glycol having an average molecular weight of 2000 to 15000 Daltons; and exposing the mixture to a pressure of 1500 to 2000 bar for a period of between 30 seconds to 10 minutes, wherein the aqueous phase makes up at least 90 volume% and the oil phase makes up at least 10 volume%. |

| Ref. | Claim |
|---|---|
| A8.13 | Method of producing a nanoemulsion, wherein the method comprises: mixing an aqueous phase with an oil phase comprising lecithin, the aqueous phase containing a botulinum protein conjugated to polyethylene glycol having an average molecular weight of 2000 to 15000 Daltons; and exposing the mixture to a pressure of 1500 to 2000 bar for a period of between 30 seconds to 10 minutes, wherein the aqueous phase makes up at least 90 volume%, the oil phase makes up at least 10 volume%, and the weight ratio of surfactant to oil is 2:1. |

## A.9. Series of claims to optical sensing device for nanosequencing

| Ref. | Claim |
|---|---|
| A9.1 | Optical sensing device. |
| A9.2 | Optical sensing device having a layer structure. |
| A9.3 | Optical sensing device having a layer structure, the layer structure comprising a substrate. |
| A9.4 | Optical sensing device having a layer structure, the layer structure comprising a substrate structured to laterally delimit a cavity. |
| A9.5 | Optical sensing device having a layer structure, the layer structure comprising: a substrate structured to laterally delimit a cavity; and a dielectric layer. |
| A9.6 | Optical sensing device having a layer structure, the layer structure comprising: a substrate structured to laterally delimit a cavity; and a dielectric layer, which extends on top of the substrate. |
| A9.7 | Optical sensing device having a layer structure, the layer structure comprising: a substrate structured to laterally delimit a cavity; and a dielectric layer, which extends on top of the substrate and forms a membrane spanning the cavity. |
| A9.8 | Optical sensing device having a layer structure, the layer structure comprising: a substrate structured to laterally delimit a cavity; and a dielectric layer, which extends on top of the substrate and forms a membrane spanning the cavity, the membrane including n apertures to the cavity, where $n \geq 1$. |
| A9.9 | Optical sensing device having a layer structure, the layer structure comprising: a substrate structured to laterally delimit a cavity; a dielectric layer, which extends on top of the substrate and forms a membrane spanning the cavity, the membrane including n apertures to the cavity, where $n \geq 1$; and n pairs of opposite antenna elements. |
| A9.10 | Optical sensing device having a layer structure, the layer structure comprising: a substrate structured to laterally delimit a cavity; a dielectric layer, which extends on top of the substrate and forms a membrane spanning the cavity, the membrane including n apertures to the cavity, where $n \geq 1$; and n pairs of opposite antenna elements, which are patterned on top of the dielectric layer. |
| A9.11 | Optical sensing device having a layer structure, the layer structure comprising: a substrate structured to laterally delimit a cavity; a dielectric layer, which extends on top of the substrate and forms a membrane spanning the cavity, the membrane including n apertures to the cavity, where $n \geq 1$; and n pairs of opposite antenna elements, which are patterned on top of the dielectric layer, on opposite lateral sides of respective ones of the n apertures. |
| A9.12 | Optical sensing device having a layer structure, the layer structure comprising: a substrate structured to laterally delimit a cavity; a dielectric layer, which extends on top of the substrate and forms a membrane spanning the cavity, the membrane including n apertures to the cavity, where $n \geq 1$; and n pairs of opposite antenna elements, which are patterned on top of the dielectric layer, on opposite lateral sides of respective ones of the n apertures, and define n respective gaps extending between opposite antenna elements of the n pairs along respective directions. |
| A9.13 | Optical sensing device having a layer structure, the layer structure comprising: a substrate structured to laterally delimit a cavity; a dielectric layer, which extends on top of the substrate and forms a membrane spanning the cavity, the membrane including n apertures to the cavity, where $n \geq 1$; and n pairs of opposite antenna elements, which are patterned on top of the dielectric layer, on opposite lateral sides of respective ones of the n apertures, and define n respective gaps extending between opposite antenna elements of the n pairs along respective directions parallel to a main plane of the substrate, so as to define n molecular passages. |
| A9.14 | Optical sensing device having a layer structure, the layer structure comprising: a substrate structured to laterally delimit a cavity; a dielectric layer, which extends on top of the substrate and forms a membrane spanning the cavity, the membrane including n apertures to the cavity, where $n \geq 1$; and n pairs of opposite antenna elements, which are patterned on top of the dielectric layer, on opposite lateral sides of respective ones of the n apertures, and define n respective gaps extending between opposite antenna elements of the n pairs along respective directions parallel to a main plane of the substrate, so as to define n molecular passages, each extending from the cavity through a respective one of the n apertures and a respective one of the n gaps along a direction transverse to the main plane of the substrate. |
| A9.15 | Optical sensing device having a layer structure, the layer structure comprising: a substrate structured to laterally delimit a cavity; a dielectric layer, which extends on top of the substrate and forms a membrane spanning the cavity, the membrane including n apertures to the cavity, where $n \geq 1$; and n pairs of opposite antenna elements, which are patterned on top of the dielectric layer, on opposite lateral sides of respective ones of the n apertures, and define n respective gaps extending between opposite antenna elements of the n pairs along respective directions parallel to a main plane of the substrate, so as to define n molecular passages, each extending from the cavity through a respective one of the n apertures and a respective one of the n gaps along a direction transverse to the main plane of the substrate, wherein an average length of the n gaps along said respective directions is between 4 nm and 20 nm. |
| A9.16 | Optical sensing device having a layer structure, the layer structure comprising: a substrate structured to laterally delimit a cavity; a dielectric layer, which extends on top of the substrate and forms a membrane spanning the cavity, the membrane including n apertures to the cavity, where $n \geq 1$; and n pairs of opposite antenna elements, which are patterned on top of the dielectric layer, on opposite lateral sides of respective ones of the n apertures, and define n respective gaps extending between opposite antenna elements of the n pairs along respective directions parallel to a main plane of the substrate, so as to define n molecular passages, each extending from the cavity through a respective one of the n apertures and a respective one of the n gaps along a direction transverse to the main plane of the substrate, wherein an average length of the n gaps along said respective directions is between 4 nm and 20 nm, whereby the n gaps define respective electromagnetic field enhancement regions. |
| A9.17 | Optical sensing device having a layer structure, the layer structure comprising: a substrate structured to laterally delimit a cavity; a dielectric layer, which extends on top of the substrate and forms a membrane spanning the cavity, the membrane including n apertures to |

| Ref. | Claim |
|---|---|
| | the cavity, where n ≥ 1; and n pairs of opposite antenna elements, which are patterned on top of the dielectric layer, on opposite lateral sides of respective ones of the n apertures, and define n respective gaps extending between opposite antenna elements of the n pairs along respective directions parallel to a main plane of the substrate, so as to define n molecular passages, each extending from the cavity through a respective one of the n apertures and a respective one of the n gaps along a direction transverse to the main plane of the substrate, wherein an average length of the n gaps along said respective directions is between 4 nm and 20 nm, whereby the n gaps define respective electromagnetic field enhancement regions, in which electromagnetic radiation can be concentrated upon irradiating the antenna elements, for optically sensing molecules, in operation. |
| A9.18 | Optical sensing device having a layer structure, the layer structure comprising: a substrate structured to laterally delimit a cavity; a dielectric layer, which extends on top of the substrate and forms a membrane spanning the cavity, the membrane including n apertures to the cavity, where n ≥ 1; and n pairs of opposite antenna elements, which are patterned on top of the dielectric layer, on opposite lateral sides of respective ones of the n apertures, and define n respective gaps extending between opposite antenna elements of the n pairs along respective directions parallel to a main plane of the substrate, so as to define n molecular passages, each extending from the cavity through a respective one of the n apertures and a respective one of the n gaps along a direction transverse to the main plane of the substrate, wherein an average length of the n gaps along said respective directions is between 4 nm and 20 nm, whereby the n gaps define respective electromagnetic field enhancement regions, in which electromagnetic radiation can be concentrated upon irradiating the antenna elements, for optically sensing molecules, in operation, and wherein an average diameter of the n apertures is larger than or equal to said average length of the gaps along said respective directions. |
| A9.19 | Optical sensing device having a layer structure, the layer structure comprising: a substrate structured to laterally delimit a cavity; a dielectric layer, which extends on top of the substrate and forms a membrane spanning the cavity, the membrane including n apertures to the cavity, where n ≥ 1; and n pairs of opposite antenna elements, which are patterned on top of the dielectric layer, on opposite lateral sides of respective ones of the n apertures, and define n respective gaps extending between opposite antenna elements of the n pairs along respective directions parallel to a main plane of the substrate, so as to define n molecular passages, each extending from the cavity through a respective one of the n apertures and a respective one of the n gaps along a direction transverse to the main plane of the substrate, wherein an average length of the n gaps along said respective directions is between 4 nm and 20 nm, whereby the n gaps define respective electromagnetic field enhancement regions, in which electromagnetic radiation can be concentrated upon irradiating the antenna elements, for optically sensing molecules, in operation, and wherein an average diameter of the n apertures is larger than or equal to said average length of the gaps along said respective directions, whereby a minimal cross-sectional dimension of each of n the passages is limited by a respective one of the n gaps along said respective directions. |
| A9.20 | Optical sensing device having a layer structure, the layer structure comprising: a substrate structured to laterally delimit a cavity; a dielectric layer, which extends on top of the substrate and forms a membrane spanning the cavity, the membrane including n apertures to the cavity, where n ≥ 100; and n pairs of opposite antenna elements, which are patterned on top of the dielectric layer, on opposite lateral sides of respective ones of the n apertures, and define n respective gaps extending between opposite antenna elements of the n pairs along respective directions parallel to a main plane of the substrate, so as to define n molecular passages, each extending from the cavity through a respective one of the n apertures and a respective one of the n gaps along a direction transverse to the main plane of the substrate, wherein an average length of the n gaps along said respective directions is between 4 nm and 20 nm, whereby the n gaps define respective electromagnetic field enhancement regions, in which electromagnetic radiation can be concentrated upon irradiating the antenna elements, for optically sensing molecules, in operation, and wherein an average diameter of the n apertures is larger than or equal to said average length of the gaps along said respective directions, whereby a minimal cross-sectional dimension of each of n the passages is limited by a respective one of the n gaps along said respective directions. |
| A9.21 | Optical sensing device having a layer structure, the layer structure comprising: a substrate structured to laterally delimit a cavity; a dielectric layer, which extends on top of the substrate and forms a membrane spanning the cavity, the membrane including n apertures to the cavity, where n ≥ 100; and n pairs of opposite antenna elements, which are patterned on top of the dielectric layer, on opposite lateral sides of respective ones of the n apertures, and define n respective gaps extending between opposite antenna elements of the n pairs along respective directions parallel to a main plane of the substrate, so as to define n molecular passages, each extending from the cavity through a respective one of the n apertures and a respective one of the n gaps along a direction transverse to the main plane of the substrate, wherein an average length of the n gaps along said respective directions is between 4 nm and 20 nm, whereby the n gaps define respective electromagnetic field enhancement regions, in which electromagnetic radiation can be concentrated upon irradiating the antenna elements, for optically sensing molecules, in operation, and wherein an average diameter of the n apertures is equal to said average length of the gaps along said respective directions, subject to a tolerance of 2 nm, whereby a minimal cross-sectional dimension of each of n the passages is limited by a respective one of the n gaps along said respective directions. |
| A9.22 | Optical sensing device having a layer structure, the layer structure comprising: a substrate structured to laterally delimit a cavity; a dielectric layer, which extends on top of the substrate and forms a membrane spanning the cavity, the membrane including n apertures to the cavity, where n ≥ 100; and n pairs of opposite antenna elements, which are patterned on top of the dielectric layer, on opposite lateral sides of respective ones of the n apertures, and define n respective gaps extending between opposite antenna elements of the n pairs along respective directions parallel to a main plane of the substrate, so as to define n molecular passages, each extending from the cavity through a respective one of the n apertures and a respective one of the n gaps along a direction transverse to the main plane of the substrate, wherein an average length of the n gaps along said respective directions is between 4 nm and 20 nm, whereby the n gaps define respective electromagnetic field enhancement regions, in which electromagnetic radiation can be concentrated upon irradiating the antenna elements, for optically sensing molecules, in operation, and wherein an average diameter of the n apertures is equal to said average length of the gaps along said respective directions, subject to a tolerance of 2 nm, whereby a minimal cross-sectional dimension of each of n the passages is limited by a respective one of the n gaps along said respective directions, and the diameters of the apertures and the lengths of the gaps are essentially constant, subject to a dispersion of less than 2 nm. |

## Appendix B: Scope results

The following provides the list of scope values obtained for claims listed in Appendix A. The first column is the claim reference. The second column shows the word count of this claim. The next seven columns list scope values obtained from the models described in section 3.

| | | Scope | | | | | | |
|---|---|---|---|---|---|---|---|---|
| Claim ref. | Word count | Davinci-002 | Babbage-002 | GPT2 | Word frequency | Word count | Character frequency | Character count |
| A1.1 | 2 | 0.053362 | 0.051220 | 0.049554 | 0.051504 | 0.041829 | 0.017298 | 0.013454 |
| A1.2 | 5 | 0.029785 | 0.029338 | 0.024945 | 0.021539 | 0.016732 | 0.008480 | 0.006554 |
| A1.3 | 8 | 0.025826 | 0.025012 | 0.015853 | 0.014016 | 0.010457 | 0.006464 | 0.004823 |
| A1.4 | 16 | 0.020386 | 0.019655 | 0.010059 | 0.007448 | 0.005229 | 0.003528 | 0.002582 |
| A1.5 | 19 | 0.016766 | 0.015826 | 0.008006 | 0.006201 | 0.004403 | 0.002958 | 0.002185 |
| A1.6 | 24 | 0.012822 | 0.012177 | 0.006708 | 0.004995 | 0.003486 | 0.002411 | 0.001788 |
| A1.7 | 26 | 0.010925 | 0.010053 | 0.006120 | 0.004497 | 0.003218 | 0.002087 | 0.001568 |
| A1.8 | 35 | 0.008926 | 0.008462 | 0.004547 | 0.003365 | 0.002390 | 0.001562 | 0.001173 |
| A1.9 | 40 | 0.008487 | 0.007923 | 0.004301 | 0.002966 | 0.002091 | 0.001390 | 0.001039 |
| A1.10 | 54 | 0.007217 | 0.006639 | 0.003479 | 0.002258 | 0.001549 | 0.001028 | 0.000772 |
| A1.11 | 61 | 0.006355 | 0.005843 | 0.003044 | 0.001986 | 0.001371 | 0.000881 | 0.000666 |
| A1.12 | 66 | 0.005514 | 0.005089 | 0.002713 | 0.001808 | 0.001268 | 0.000799 | 0.000606 |
| A1.13 | 70 | 0.004940 | 0.004450 | 0.002433 | 0.001679 | 0.001195 | 0.000755 | 0.000572 |
| A1.14 | 81 | 0.004286 | 0.003852 | 0.002202 | 0.001454 | 0.001033 | 0.000657 | 0.000497 |
| A1.15 | 94 | 0.003966 | 0.003474 | 0.001902 | 0.001260 | 0.000890 | 0.000572 | 0.000434 |
| A1.16 | 118 | 0.003255 | 0.002812 | 0.001613 | 0.001044 | 0.000709 | 0.000468 | 0.000367 |
| A1.17 | 127 | 0.003009 | 0.002610 | 0.001505 | 0.000956 | 0.000659 | 0.000429 | 0.000336 |
| A1.18 | 131 | 0.002953 | 0.002565 | 0.001413 | 0.000923 | 0.000639 | 0.000415 | 0.000324 |
| A1.19 | 134 | 0.002913 | 0.002516 | 0.001338 | 0.000894 | 0.000624 | 0.000403 | 0.000316 |
| A1.20 | 147 | 0.002611 | 0.002247 | 0.001226 | 0.000816 | 0.000569 | 0.000364 | 0.000291 |
| A1.21 | 153 | 0.002493 | 0.002151 | 0.001153 | 0.000779 | 0.000547 | 0.000347 | 0.000277 |
| A1.22 | 163 | 0.002409 | 0.002075 | 0.001094 | 0.000732 | 0.000513 | 0.000328 | 0.000264 |
| A2.1 | 10 | 0.019883 | 0.019548 | 0.016235 | 0.010522 | 0.008366 | 0.004705 | 0.003600 |
| A2.2 | 18 | 0.014741 | 0.013660 | 0.011145 | 0.006495 | 0.004648 | 0.002822 | 0.002148 |
| A2.3 | 27 | 0.011219 | 0.009792 | 0.007612 | 0.004311 | 0.003098 | 0.001803 | 0.001360 |
| A2.4 | 36 | 0.009019 | 0.007899 | 0.005336 | 0.003181 | 0.002324 | 0.001316 | 0.001006 |
| A2.5 | 50 | 0.007772 | 0.006850 | 0.004609 | 0.002394 | 0.001673 | 0.001009 | 0.000765 |
| A2.6 | 62 | 0.006757 | 0.006055 | 0.004105 | 0.002012 | 0.001349 | 0.000852 | 0.000644 |
| A2.7 | 64 | 0.006428 | 0.005762 | 0.003917 | 0.001929 | 0.001307 | 0.000802 | 0.000607 |
| A2.8 | 72 | 0.006307 | 0.005648 | 0.003794 | 0.001719 | 0.001162 | 0.000727 | 0.000549 |
| A2.9 | 78 | 0.005774 | 0.005103 | 0.003429 | 0.001575 | 0.001073 | 0.000663 | 0.000500 |
| A2.10 | 81 | 0.005481 | 0.004785 | 0.003130 | 0.001509 | 0.001033 | 0.000635 | 0.000481 |
| A2.11 | 92 | 0.005177 | 0.004445 | 0.002812 | 0.001322 | 0.000909 | 0.000567 | 0.000428 |
| A2.12 | 132 | 0.003982 | 0.003417 | 0.002128 | 0.000944 | 0.000634 | 0.000422 | 0.000316 |
| A2.13 | 155 | 0.003837 | 0.003299 | 0.001978 | 0.000814 | 0.000540 | 0.000368 | 0.000276 |
| A2.14 | 180 | 0.003327 | 0.002878 | 0.001755 | 0.000714 | 0.000465 | 0.000322 | 0.000241 |
| A2.15 | 209 | 0.002820 | 0.002398 | 0.001505 | 0.000617 | 0.000400 | 0.000280 | 0.000210 |
| A2.16 | 247 | 0.002546 | 0.002177 | 0.001365 | 0.000527 | 0.000339 | 0.000239 | 0.000179 |
| A2.17 | 283 | 0.002092 | 0.001829 | 0.001185 | 0.000461 | 0.000296 | 0.000211 | 0.000158 |
| A2.18 | 298 | 0.001971 | 0.001724 | 0.001133 | 0.000437 | 0.000281 | 0.000200 | 0.000149 |
| A2.19 | 320 | 0.001842 | 0.001600 | 0.001027 | 0.000405 | 0.000261 | 0.000185 | 0.000138 |

| Claim ref. | Word count | Scope: Davinci-002 | Babbage-002 | GPT2 | Word frequency | Word count | Character frequency | Character count |
|---|---|---|---|---|---|---|---|---|
| A3.1 | 2 | 0.059404 | 0.061953 | 0.052748 | 0.047554 | 0.041829 | 0.022057 | 0.017041 |
| A3.2 | 6 | 0.032375 | 0.030483 | 0.024026 | 0.018110 | 0.013943 | 0.007536 | 0.005810 |
| A3.3 | 13 | 0.015198 | 0.015293 | 0.010239 | 0.008099 | 0.006435 | 0.003236 | 0.002531 |
| A3.4 | 20 | 0.012536 | 0.012463 | 0.006475 | 0.005393 | 0.004183 | 0.002164 | 0.001693 |
| A3.5 | 25 | 0.010081 | 0.010024 | 0.005385 | 0.004447 | 0.003346 | 0.001789 | 0.001397 |
| A3.6 | 30 | 0.008883 | 0.008514 | 0.004474 | 0.003801 | 0.002789 | 0.001524 | 0.001183 |
| A3.7 | 38 | 0.008023 | 0.007408 | 0.003538 | 0.003009 | 0.002202 | 0.001216 | 0.000954 |
| A3.8 | 50 | 0.006911 | 0.006266 | 0.002679 | 0.002255 | 0.001673 | 0.000938 | 0.000732 |
| A3.9 | 63 | 0.006001 | 0.005348 | 0.002259 | 0.001828 | 0.001328 | 0.000750 | 0.000588 |
| A3.10 | 84 | 0.004917 | 0.004318 | 0.001719 | 0.001390 | 0.000996 | 0.000567 | 0.000443 |
| A3.11 | 104 | 0.004219 | 0.003629 | 0.001491 | 0.001137 | 0.000804 | 0.000465 | 0.000361 |
| A3.12 | 121 | 0.003855 | 0.003351 | 0.001306 | 0.000977 | 0.000691 | 0.000399 | 0.000309 |
| A3.13 | 140 | 0.003480 | 0.003071 | 0.001163 | 0.000858 | 0.000598 | 0.000349 | 0.000269 |
| A3.14 | 161 | 0.003257 | 0.002876 | 0.001070 | 0.000745 | 0.000520 | 0.000305 | 0.000235 |
| A3.15 | 173 | 0.003057 | 0.002699 | 0.000980 | 0.000691 | 0.000484 | 0.000284 | 0.000219 |
| | | | | | | | | |
| A4.1 | 2 | 0.045297 | 0.043015 | 0.021414 | 0.036136 | 0.041829 | 0.011507 | 0.009129 |
| A4.2 | 7 | 0.023480 | 0.022297 | 0.009398 | 0.013166 | 0.011951 | 0.004970 | 0.003933 |
| A4.3 | 14 | 0.012491 | 0.011168 | 0.004130 | 0.006338 | 0.005976 | 0.002879 | 0.002434 |
| A4.4 | 22 | 0.008161 | 0.007400 | 0.003154 | 0.004181 | 0.003803 | 0.001828 | 0.001513 |
| A4.5 | 28 | 0.007086 | 0.006446 | 0.002729 | 0.003347 | 0.002988 | 0.001408 | 0.001157 |
| A4.6 | 40 | 0.006275 | 0.005718 | 0.002287 | 0.002509 | 0.002091 | 0.001064 | 0.000867 |
| A4.7 | 46 | 0.006070 | 0.005452 | 0.002167 | 0.002225 | 0.001819 | 0.000943 | 0.000763 |
| A4.8 | 64 | 0.005174 | 0.004615 | 0.001859 | 0.001734 | 0.001307 | 0.000737 | 0.000594 |
| A4.9 | 70 | 0.004875 | 0.004333 | 0.001806 | 0.001597 | 0.001195 | 0.000688 | 0.000562 |
| A4.10 | 82 | 0.004618 | 0.004109 | 0.001455 | 0.001358 | 0.001020 | 0.000582 | 0.000476 |
| A4.11 | 88 | 0.004249 | 0.003773 | 0.001319 | 0.001251 | 0.000951 | 0.000542 | 0.000441 |
| A4.12 | 84 | 0.004234 | 0.003788 | 0.001392 | 0.001319 | 0.000996 | 0.000567 | 0.000464 |
| A4.13 | 84 | 0.004267 | 0.003793 | 0.001394 | 0.001319 | 0.000996 | 0.000567 | 0.000463 |
| A4.14 | 84 | 0.004227 | 0.003791 | 0.001390 | 0.001319 | 0.000996 | 0.000569 | 0.000465 |
| | | | | | | | | |
| A5.1 | 12 | 0.023232 | 0.020738 | 0.013597 | 0.010246 | 0.006972 | 0.003617 | 0.002872 |
| A5.2 | 22 | 0.018225 | 0.016195 | 0.009329 | 0.005829 | 0.003803 | 0.002278 | 0.001751 |
| A5.3 | 40 | 0.010549 | 0.009850 | 0.005587 | 0.003290 | 0.002091 | 0.001367 | 0.001027 |
| A5.4 | 48 | 0.008383 | 0.007503 | 0.004403 | 0.002703 | 0.001743 | 0.001098 | 0.000827 |
| A5.5 | 67 | 0.005692 | 0.005147 | 0.003396 | 0.001906 | 0.001249 | 0.000802 | 0.000603 |
| A5.6 | 109 | 0.003891 | 0.003472 | 0.002290 | 0.001228 | 0.000768 | 0.000513 | 0.000384 |
| A5.7 | 137 | 0.003192 | 0.002849 | 0.001894 | 0.000979 | 0.000611 | 0.000409 | 0.000305 |
| | | | | | | | | |
| A6.1 | 7 | 0.024161 | 0.025449 | 0.022386 | 0.017095 | 0.011951 | 0.007662 | 0.005810 |
| A6.2 | 13 | 0.018825 | 0.019235 | 0.015069 | 0.009680 | 0.006435 | 0.003916 | 0.002972 |
| A6.3 | 26 | 0.011953 | 0.011471 | 0.008459 | 0.005097 | 0.003218 | 0.002162 | 0.001649 |
| A6.4 | 54 | 0.008629 | 0.008020 | 0.004603 | 0.002371 | 0.001549 | 0.001047 | 0.000791 |
| A6.5 | 64 | 0.007327 | 0.006889 | 0.003707 | 0.001983 | 0.001307 | 0.000865 | 0.000655 |
| A6.6 | 96 | 0.005431 | 0.005024 | 0.002690 | 0.001356 | 0.000871 | 0.000592 | 0.000448 |
| A6.7 | 128 | 0.004638 | 0.004422 | 0.002332 | 0.001023 | 0.000654 | 0.000449 | 0.000339 |
| A6.8 | 148 | 0.003963 | 0.003832 | 0.001902 | 0.000886 | 0.000565 | 0.000391 | 0.000295 |
| A6.9 | 169 | 0.003356 | 0.003166 | 0.001545 | 0.000772 | 0.000495 | 0.000342 | 0.000259 |
| A6.10 | 190 | 0.002994 | 0.002764 | 0.001418 | 0.000691 | 0.000440 | 0.000306 | 0.000231 |

| Claim ref. | Word count | Scope<br>Davinci-002 | Babbage-002 | GPT2 | Word frequency | Word count | Character frequency | Character count |
|---|---|---|---|---|---|---|---|---|
| A7.1 | 2 | 0.055222 | 0.052580 | 0.050031 | 0.048338 | 0.041829 | 0.027212 | 0.019663 |
| A7.2 | 9 | 0.023526 | 0.021354 | 0.018642 | 0.012001 | 0.009295 | 0.005328 | 0.003994 |
| A7.3 | 19 | 0.014190 | 0.013061 | 0.010351 | 0.005875 | 0.004403 | 0.002609 | 0.001982 |
| A7.4 | 26 | 0.012769 | 0.011454 | 0.008721 | 0.004420 | 0.003218 | 0.001972 | 0.001513 |
| A7.5 | 64 | 0.007144 | 0.006149 | 0.004315 | 0.001824 | 0.001307 | 0.000814 | 0.000622 |
| A7.6 | 79 | 0.005870 | 0.005127 | 0.003578 | 0.001451 | 0.001059 | 0.000658 | 0.000505 |
| A7.7 | 105 | 0.004779 | 0.004102 | 0.002857 | 0.001126 | 0.000797 | 0.000501 | 0.000382 |
| A7.8 | 122 | 0.004096 | 0.003531 | 0.002501 | 0.000962 | 0.000686 | 0.000427 | 0.000326 |
| A7.9 | 144 | 0.003684 | 0.003217 | 0.002216 | 0.000816 | 0.000581 | 0.000357 | 0.000272 |
| A7.10 | 162 | 0.003325 | 0.002898 | 0.002022 | 0.000732 | 0.000516 | 0.000315 | 0.000239 |
| A7.11 | 175 | 0.003121 | 0.002737 | 0.001904 | 0.000687 | 0.000478 | 0.000296 | 0.000225 |
| A7.12 | 188 | 0.002855 | 0.002520 | 0.001792 | 0.000645 | 0.000445 | 0.000276 | 0.000209 |
| A7.13 | 214 | 0.002573 | 0.002255 | 0.001625 | 0.000575 | 0.000391 | 0.000246 | 0.000187 |
| A7.14 | 257 | 0.002159 | 0.001911 | 0.001354 | 0.000481 | 0.000326 | 0.000208 | 0.000158 |
| | | | | | | | | |
| A8.1 | 5 | 0.034507 | 0.033959 | 0.025539 | 0.022733 | 0.016732 | 0.009466 | 0.007303 |
| A8.2 | 17 | 0.020739 | 0.018746 | 0.009535 | 0.006810 | 0.004921 | 0.003097 | 0.002389 |
| A8.3 | 24 | 0.013300 | 0.012185 | 0.006140 | 0.004765 | 0.003486 | 0.002071 | 0.001618 |
| A8.4 | 28 | 0.011008 | 0.010302 | 0.004249 | 0.003944 | 0.002988 | 0.001687 | 0.001331 |
| A8.5 | 38 | 0.009163 | 0.008641 | 0.003204 | 0.002907 | 0.002202 | 0.001239 | 0.001014 |
| A8.6 | 50 | 0.007764 | 0.007206 | 0.002891 | 0.002332 | 0.001673 | 0.000997 | 0.000817 |
| A8.7 | 60 | 0.006941 | 0.006510 | 0.002712 | 0.002010 | 0.001394 | 0.000864 | 0.000706 |
| A8.8 | 60 | 0.006716 | 0.006330 | 0.002703 | 0.002008 | 0.001394 | 0.000853 | 0.000708 |
| A8.9 | 63 | 0.006433 | 0.006023 | 0.002504 | 0.001899 | 0.001328 | 0.000803 | 0.000664 |
| A8.10 | 62 | 0.006312 | 0.005816 | 0.002502 | 0.001911 | 0.001349 | 0.000811 | 0.000671 |
| A8.11 | 73 | 0.005337 | 0.004938 | 0.002154 | 0.001624 | 0.001146 | 0.000704 | 0.000585 |
| A8.12 | 84 | 0.005133 | 0.004826 | 0.002086 | 0.001436 | 0.000996 | 0.000637 | 0.000528 |
| A8.13 | 93 | 0.004611 | 0.004356 | 0.001834 | 0.001304 | 0.000900 | 0.000582 | 0.000481 |
| | | | | | | | | |
| A9.1 | 3 | 0.048852 | 0.042363 | 0.038693 | 0.030108 | 0.027886 | 0.014595 | 0.011114 |
| A9.2 | 7 | 0.028465 | 0.026496 | 0.020917 | 0.015457 | 0.011951 | 0.006949 | 0.005325 |
| A9.3 | 13 | 0.021300 | 0.020072 | 0.013693 | 0.008820 | 0.006435 | 0.003621 | 0.002779 |
| A9.4 | 19 | 0.012761 | 0.012291 | 0.007032 | 0.005814 | 0.004403 | 0.002512 | 0.001922 |
| A9.5 | 23 | 0.011498 | 0.011290 | 0.005718 | 0.004941 | 0.003637 | 0.002089 | 0.001618 |
| A9.6 | 30 | 0.009530 | 0.009254 | 0.005133 | 0.003975 | 0.002789 | 0.001680 | 0.001298 |
| A9.7 | 37 | 0.008204 | 0.007932 | 0.004564 | 0.003259 | 0.002261 | 0.001395 | 0.001074 |
| A9.8 | 49 | 0.006119 | 0.005905 | 0.003424 | 0.002463 | 0.001707 | 0.001094 | 0.000849 |
| A9.9 | 55 | 0.005274 | 0.005023 | 0.003058 | 0.002171 | 0.001521 | 0.000976 | 0.000754 |
| A9.10 | 64 | 0.004967 | 0.004639 | 0.002638 | 0.001892 | 0.001307 | 0.000851 | 0.000654 |
| A9.11 | 75 | 0.004524 | 0.004172 | 0.002228 | 0.001621 | 0.001115 | 0.000735 | 0.000561 |
| A9.12 | 92 | 0.003724 | 0.003410 | 0.001888 | 0.001303 | 0.000909 | 0.000588 | 0.000446 |
| A9.13 | 107 | 0.003211 | 0.002950 | 0.001694 | 0.001136 | 0.000782 | 0.000518 | 0.000391 |
| A9.14 | 139 | 0.002911 | 0.002623 | 0.001435 | 0.000905 | 0.000602 | 0.000411 | 0.000309 |
| A9.15 | 158 | 0.002636 | 0.002343 | 0.001327 | 0.000794 | 0.000529 | 0.000366 | 0.000275 |
| A9.16 | 168 | 0.002455 | 0.002152 | 0.001240 | 0.000736 | 0.000498 | 0.000336 | 0.000254 |
| A9.17 | 186 | 0.002140 | 0.001866 | 0.001088 | 0.000660 | 0.000450 | 0.000295 | 0.000222 |
| A9.18 | 211 | 0.002000 | 0.001738 | 0.001011 | 0.000587 | 0.000396 | 0.000263 | 0.000198 |
| A9.19 | 237 | 0.001784 | 0.001553 | 0.000922 | 0.000525 | 0.000353 | 0.000236 | 0.000177 |
| A9.20 | 237 | 0.001763 | 0.001535 | 0.000919 | 0.000524 | 0.000353 | 0.000235 | 0.000177 |
| A9.21 | 241 | 0.001708 | 0.001494 | 0.000903 | 0.000517 | 0.000347 | 0.000232 | 0.000175 |
| A9.22 | 265 | 0.001568 | 0.001379 | 0.000825 | 0.000476 | 0.000316 | 0.000213 | 0.000160 |

# Appendix C – Additional claim sets

Section C.1 lists claim variations on a pencil eraser, where each claim is 25-word long. Section C.2 shows the sets of lists considered to build claim variations by combining alternative elements of the lists as tuples.

## C.1. Claim variations (each 25-word long) on a pencil eraser

| Ref. | Claim |
|---|---|
| C1.1 | Writing instrument, wherein the instrument comprises a pencil, an eraser, and a ferrule, and wherein the eraser is attached to the pencil through the ferrule |
| C1.2 | Writing instrument, wherein the instrument comprises a cedarwood pencil, an eraser, and a ferrule, wherein the eraser is attached to the pencil through the ferrule |
| C1.3 | Writing instrument, which comprises a cedarwood pencil, an eraser, and a metal ferrule, in which the eraser is attached to the pencil through the ferrule |
| C1.4 | Writing instrument comprising a cedarwood pencil, an eraser, and a metal ferrule having serrations, wherein the eraser is attached to the pencil through the ferrule |
| C1.5 | Writing instrument comprising a cedarwood pencil, an eraser, and an aluminum ferrule having serrations, wherein the eraser is attached to the pencil through the ferrule |
| C1.6 | Writing instrument comprising a cedarwood pencil, an aluminum ferrule having longitudinal serrations, and an eraser, the eraser being attached to the pencil through the ferrule |
| C1.7 | Writing instrument comprising a cedarwood pencil, an aluminum ferrule having longitudinal serrations, and an eraser being removably attached to the cedarwood pencil through the ferrule |
| C1.8 | Writing instrument comprising a western redcedar wood pencil, an aluminum ferrule having longitudinal serrations, and an eraser removably attached to the pencil through the ferrule |

## C.2. Sets of lists used to build claim variations

### C.2.1. Pencil eraser

{{"writing instrument"}, {"comprising", "including", ", which comprises", ", which includes", ", the instrument comprising", ", wherein the instrument comprises",", wherein the writing instrument comprises", "having"}, {"a pencil, a rubber, and a ferrule"}, {", wherein", ", in which"}, {"the rubber is attached to the pencil", "the rubber is fixed to the pencil"}, {"through", "via", "by means of"}, {"the ferrule"}}

## C.2.2. Optical device

{{"optical device"}, {"comprising", "including", ", which comprises", ", which includes",", the device comprising",", the optical device comprising"}, {"a layer structure"}, {", the layer structure including", ", which comprises", ", which includes", "including"}, {": a thermally conducting optical reflector; a thermally conducting spacer"}, {"arranged above the reflector", ", which is arranged above the reflector"}, {"; and a phase change material"}, {"arranged above the spacer", ", which is arranged above the spacer"}, {", wherein the phase change material has at least two reversibly switchable states, in which the phase change material exhibits two different values of refractive index"}}

## C.2.3. Biodegradable nanoparticles

{{"Biodegradable nanoparticles"}, {", wherein the nanoparticles are",", the nanoparticles being ", " "}, {"based on a biodegradable homopolymer of C2–C6 alkyl 2-cyanoacrylate monomers"}, {", which are", " "}, {"selected from ethyl 2-cyanoacrylate and n-butyl 2-cyanoacrylate"}, {", wherein", ", in which"}, {"the nanoparticles"}, {"comprise", "include"}, {"a non-covalent insulin complex"}, {" ", ", which is"}, {"encapsulated therein", "encapsulated in it"}, {"along with", "together with", "with"}, {"a pharmaceutically acceptable stabilizer"}}

## C.2.4. Datacentric processing

{{"method"}, {"of"}, {"managing processing resources of a computerized system"}, {"having", "including", "comprising", ", which includes", ", which comprises",", the computerized system comprising",", the system comprising",", wherein the computerized system comprises",", wherein the system comprises",", wherein the computerized system includes",", wherein the system includes"}, {"a first processor and a second processor, "}, {"the method comprising: ","said method comprising: ", " wherein the method comprises: "}, {"monitoring data processed by the first processor while executing; and, if the second processor is found", "monitoring data being processed by the first processor while executing; and, if the second processor is found"}, {"to be at", "at"}, {"a shorter distance than the first processor from the monitored data, instructing to interrupt an execution at the first processor and resume the execution at the second processor"}}

## C.2.5. Nanoemulsion

{{"method"}, {"of producing a nanoemulsion"}, {", wherein the method comprises: ", ", the method comprising: ",", comprising: "}, {"mixing an aqueous phase with an oil phase"}, {"comprising lecithin", "including lecithin"}, {", the aqueous phase comprising",", the aqueous phase including",", wherein the aqueous phase comprises",", wherein the aqueous phase includes"}, {"a botulinum protein"}, {", which is", " "}, {"conjugated to polyethylene glycol"}, {"having"}, {"an average molecular weight of"}, {"between", " "}, {"2000 to 15000 Daltons; and exposing the mixture to a pressure of 1500 to 2000 bar for a period of"}, {"between", " "}, {"30 seconds to 10 minutes"}}

---